%% file: acl_latex.tex
\documentclass[11pt]{article}

\usepackage[final]{acl}

\usepackage{times}
\usepackage{latexsym}
\usepackage{amsmath}
\usepackage{amssymb}
\usepackage{booktabs}
\usepackage{multirow}
\usepackage{makecell}
\usepackage{colortbl}
\usepackage[table]{xcolor}
\usepackage{arydshln}     
\usepackage{subcaption,tcolorbox}

\definecolor{oursrow}{RGB}{230,241,251}   
\definecolor{bestval}{RGB}{12,68,124}     
\newcommand{\best}[1]{\textcolor{black}{\textbf{#1}}}
\newcommand{\ours}{\rowcolor{oursrow}}
\newcommand{\ph}{--}                       

\usepackage[T1]{fontenc}
\usepackage[utf8]{inputenc}
\usepackage{microtype}
\usepackage{inconsolata}
\usepackage{graphicx}

\usepackage{appendix}

\usepackage{soul}
\usepackage{cuted}           

\newcommand{\sys}{\textsc{MoCA-Agent}\xspace}
\usepackage{xspace}

\title{MoCA-Agent: A Market-of-Claims Code Agent for Financial and Numerical Reasoning}

\author{
  \textbf{Abdelrahman Abdallah\textsuperscript{1}, AbdelRahim A. Elmadany\textsuperscript{2}, Sameh Al Natour\textsuperscript{3}, } \\ \textbf{Hasan Cavusoglu\textsuperscript{2}, Adam Jatowt\textsuperscript{1}, Muhammad Abdul-Mageed $^{\textsuperscript{2},\lambda}$} \\
  \textsuperscript{1}University of Innsbruck \quad \textsuperscript{2}University of British Columbia \quad \textsuperscript{3}Toronto Metropolitan University \\
  $^{\lambda}$Canada Research Chair in NLP and ML \\
  \texttt{muhammad.mageed@ubc.ca}
}

\begin{document}
\maketitle

\begin{strip}
\vspace*{-3em}
  \centering
  \includegraphics[width=\textwidth]{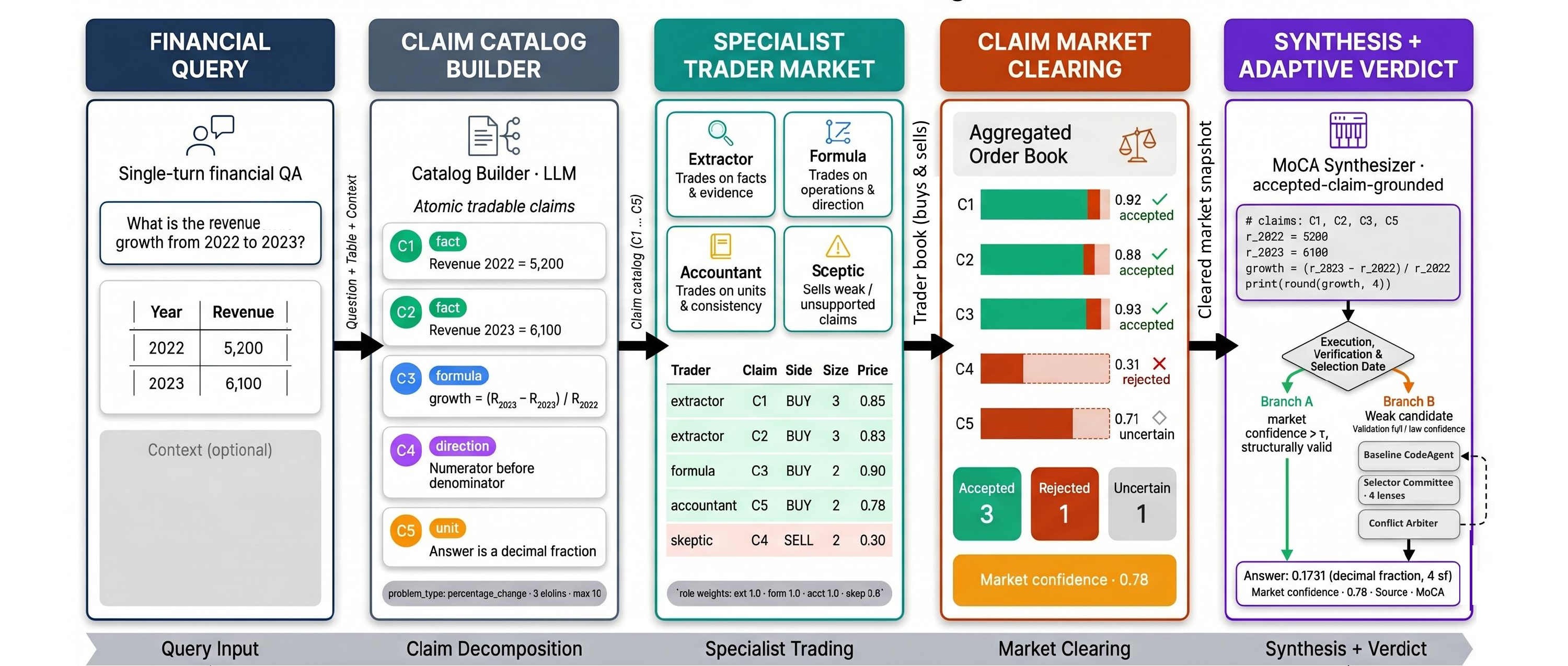}
  \captionof{figure}{Overview of \sys. A claim catalog is built from the question, table, and context. Specialist traders place buy/sell orders on claims; the market clears them into prices, confidences, and statuses. A synthesizer writes a Python program from market-supported claims; an executor runs it, and a verifier validates the output. }
    \label{fig:overview}
\end{strip}

\input{sections/abstract}
\input{sections/introduction}

\input{sections/lit}

\input{sections/our_method}
\input{sections/experiments}
\input{sections/results}
\input{sections/discussion}

\input{sections/conclusion}
\input{sections/limit}

\input{sections/ethics}



\bibliography{custom}

\appendix
\input{sections/appendix}

\end{document}

%% file: sections/abstract.tex
\begin{abstract}

Financial and tabular question answering requires more than fluent reasoning: answers must be grounded in the exact facts, formulas, units, signs, and scales that support them. A single misread cell or incorrect operation can silently produce a plausible but wrong result. We introduce \textsc{MOCA-Agent}, a market-of-claims code agent that replaces free-form multi-agent debate with claim-level verification. The system decomposes each question into typed atomic claims, asks specialist trader agents to buy or sell those claims, clears their orders into confidence-weighted accept/reject decisions, and synthesizes an executable Python program from market-supported evidence. A code-aware verifier then checks the program for execution, structural consistency, and common financial reasoning errors, with at most one market-aware repair round. Across ten public benchmarks spanning financial numerical reasoning, general tabular reasoning, ESG question answering, and multimodal chart reasoning, \textsc{MOCA-Agent} achieves strong performance using a fixed Qwen3.6-27B backbone, including $78.3\%$ on FinQA, $76.0\%$ on FinanceMath, $71.2\%$ on MultiHiertt, $86.9\%$ on ESGenius, and $85.6\%$ average on FinChart-Bench. These results show that aggregating evidence at the level of atomic claims, rather than whole answers, improves robustness in high-stakes numerical reasoning.\footnote{The code and data are available: \url{https://github.com/UBC-NLP/MoCA-Agent}.}
\end{abstract}

%% file: sections/introduction.tex
\section{Introduction}
\label{sec:introduction}

Financial and tabular question answering systems are increasingly being
deployed as the analytic backbone of automated reporting, audit, and
research-assistant pipelines~\cite{xie2023pixiu,xie2024finben,
krumdick2024bizbench}. Yet, unlike open-ended chat tasks, this regime
has virtually no tolerance for fluent-but-wrong derivations: a single misread
cell, swapped formula, or off-by-100-percent error silently
corrupts the final number while the surrounding prose continues to
read plausible~\cite{chen2021finqa,zhao2022multihiertt,zhao2024docmath,
zhao2024financemath}. Standard approaches for closing this gap, including
chain-of-thought (CoT)~\cite{wei2022chain},
program-of-thought (PoT)~\cite{chen2022pot,lu2023chameleon}, and
free-form multi-agent debate~\cite{du2023improving,madaan2023selfrefine,
shinn2023reflexion}, address part of the problem. Still, they
share a deeper limitation: they operate at the level of complete \emph{outputs} rather than
\emph{the atomic claims those outputs depend on}. 

Concretely, three failure modes arise in financial code agents: \texttt{(I) Silent Miscomputation in PoT:}
A PoT agent effectively commits once it emits a program.
Even a syntactically valid program can divide the wrong cell by another
wrong cell, while downstream consistency checks observe only the printed
output rather than the facts, units, signs, and formulas the program assumed. 
On MultiHiertt~\cite{zhao2022multihiertt}, OpenAI o1, as reported by \citet{cao2025fortune}, achieves only 38\% with textual reasoning and 49\%
with symbolic reasoning, despite the full table content being provided in the input context. \texttt{(II) Opaque Aggregation in Multi-Agent Debate:}
Both free-form debate~\cite{du2023improving} and critic-actor
loops~\cite{shinn2023reflexion,madaan2023selfrefine,gou2024critic}
typically aggregate \emph{whole answers} or whole critiques. If multiple agents fluently misderive
the same wrong number, the consensus can become wrong with high confidence.
Recent multi-agent financial pipelines~\cite{tan2025improved,
finagentcasc2025} improve robustness, but they still treat each agent's
output as an opaque vote rather than a collection of typed claims that other agents can independently support or challenge. \texttt{(III) Insufficient Structural Verification:}
Self-refine and code-repair frameworks~\cite{madaan2023selfrefine,
shinn2023reflexion,yi2025reveal,arcs2025} also use code or iterative
repair. However, their feedback is mainly based on execution failures
or free-form critique, making it difficult to detect programs
that execute successfully while using the wrong fact, formula, sign, or
scale. Common financial mistakes, such as scaling a ratio by 100,
flipping the sign on a net-of-tax line, or summing the wrong row
total, all execute cleanly. As a result, neither runtime feedback nor
free-form critique reliably surfaces the silent-error patterns that dominate financial numerical reasoning. 


This work makes four main contributions. \textbf{(1) Failure-Mode Analysis and System Design.} We identify three recurring failure modes in financial and tabular code agents: \emph{silent miscomputation}, in which an answer is numerically wrong despite appearing plausible; \emph{opaque aggregation}, where whole-agent outputs are combined without exposing the factual, formulaic, or unit-level assumptions on which they depend; and \emph{insufficient structural verification}, where generated programs are not checked against the arithmetic, accounting, and scale constraints of the task. To address these failures jointly, we propose \sys (Figure~\ref{fig:overview}), a code-agent framework that makes these assumptions explicit as typed atomic claims and aggregates them through a structured claim market rather than free-form debate (\S\ref{sec:method}).
\textbf{(2) Market-of-Claims Protocol.} In this protocol, specialist roles (\textsc{extractor}, \textsc{formula}, \textsc{accountant}, and \textsc{skeptic}) place weighted, signed orders over typed atomic claims, including facts, formulas, units, signs, and directions of change. The market clears these orders into per-claim prices, confidence scores, and accept/reject/uncertain statuses (\S\ref{sec:market}), creating an explicit audit trail of the assumptions underlying the final program. The synthesizer is restricted to non-rejected claims, and the verifier uses the same market record to determine whether the generated program is sufficiently grounded or requires repair.
\textbf{(3) Code-Aware Verification and Conflict Recovery.} We make the \textsc{Code-Aware Verifier} a core routing component rather than a passive post-hoc sanity check. After executing the synthesized program, the verifier applies operation-aware structural tests that target the most common silent-error patterns in financial reasoning, including sign flips, percent-scale errors, and missing or inconsistent formulas. When the market-grounded program is structurally weak, a \textsc{Hybrid Selector Committee} compares it with an alternative candidate; if the candidates expose unresolved disagreements, a \textsc{Conflict Arbiter} produces a grounded recovery program (\S\ref{sec:selector}). 
\textbf{(4) Benchmark Results Across Financial and Tabular Tasks.} We evaluate \sys on 10 public financial and tabular benchmarks. Using a single Qwen3.6-27B backbone and shared prompting setup, \sys matches or surpasses the strongest previously reported results in our comparison on six tracks. The largest gains appear in settings that require both multi-cell extraction and formula composition: $+14.4$ on MultiHiertt over RL-trained Fortune, $+9.0$ on FinanceMath validation over GPT-4o (PoT), $+8.3$ on DocMath-Complong over DeepSeek-V3, and $+4.1$ on FinQA over Fino1-14B (\S\ref{sec:experiments}).

%% file: sections/lit.tex
\section{Literature Review} \label{sec:related_work}

Financial agents have been studied in several forms, from trading and portfolio-management systems to risk-analysis tools and retrieval-based financial question answering. In this work, we focus on a more specific setting: code-producing agents for financial and tabular question answering. These agents must retrieve or extract relevant evidence, reason over tables and text, compose the appropriate calculation, and often generate the executable code for the final answer. We therefore review work along four dimensions: financial and table-reasoning benchmarks, program- and tool-augmented reasoning, multi-agent aggregation, and verification or repair of generated code.
\paragraph{Benchmarks and code-augmented agents.}
Financial numerical-reasoning datasets such as FinQA~\cite{chen2021finqa}, ConvFinQA~\cite{chen2022convfinqa}, TAT-QA~\cite{zhu2021tatqa}, MultiHiertt~\cite{zhao2022multihiertt}, and HiTab~\cite{cheng2022hitab} established text--table reasoning settings that require evidence extraction and program-like numerical derivations. Subsequent work extends this setting to long documents (DocFinQA~\cite{reddy2024docfinqa}, DocMath-Eval~\cite{zhao2024docmath}), harder financial numerical reasoning (FinanceMath~\cite{zhao2024financemath}, FinanceReasoning~\cite{zhu2025financereasoning}), and broader financial or business knowledge probes  (ESGenius~\cite{esgenius2025}, FinBen~\cite{xie2024finben}, PIXIU~\cite{xie2023pixiu}, BizBench~\cite{krumdick2024bizbench}); general-tabular benchmarks include WTQ~\cite{pasupat2015wtq} and TabMWP~\cite{lu2023tabmwp}. Program-of-thought~\cite{chen2022pot} and chain-of-table~\cite{wang2024chainoftable} ground reasoning in executable artifacts, while recent tabular code agents combine plan synthesis with iterative table operations (TableMaster~\cite{cao2025tablemaster}, ReAcTable~\cite{zhang2024reactable}, TIDE-Agent~\cite{deng2025tide}, ARTEMIS-DA~\cite{artemis2024}, TableMind~\cite{wang2025tablemind}). For finance and hierarchical tables, representative strong baselines include Fortune~\cite{cao2025fortune}, TableGPT2~\cite{li2024tablegpt2}, TabAF~\cite{wang2025tabaf}, and Fin-o1~\cite{liu2025fino1}, with multi-agent extensions in \citet{tan2025improved} and \citet{finagentcasc2025}. 
\sys instead exposes these assumptions as typed atomic claims before code generation, allowing specialist roles to accept, reject, or challenge the relevant fact, formula, unit, sign, or direction before the final program is synthesized.


\paragraph{Debate, verification, and code repair.}
\citet{du2023improving} show that debate among LLM instances improves
factuality, while ReAct~\cite{yao2023react} interleaves reasoning with
tool use. Recent work further formalizes deliberation~\cite{can-llms-debate2025,
mad-fact2025,mad-judge2025,mad-math2025}, but
\citet{can-llms-debate2025} note that, under matched compute, debate can
reduce to ensembling unless agents disagree over \emph{structured}
sub-claims. Self-Refine~\cite{madaan2023selfrefine},
Reflexion~\cite{shinn2023reflexion}, CRITIC~\cite{gou2024critic}, and
code-repair systems (ReVeal~\cite{yi2025reveal},
ARCS~\cite{arcs2025}) iterate on execution signals or LLM-judge
critique; these catch crashes but often miss silent financial errors
(e.g., sign-flipped net of tax, percent scaled by 100, or ratios with the
wrong denominator). In contrast, \sys makes structured sub-claims
first-class: traders can short an over-confident formula without rejecting
the rest of the program, while the verifier derives operation-specific
checks from a typed problem class and supplies both runtime and structural
feedback to a single market-aware repair round (\S\ref{sec:verifier}).

%% file: sections/our_method.tex

\section{\sys}
\label{sec:method}

In this section, we present \sys, a framework for financial and tabular question answering. Given a question \(q\), an optional table \(T\), and free-form context \(C\), \sys produces an executable Python program \(\mathcal{P}\), whose printed output is returned as the final answer \(a\). Unlike free-form multi-agent debate systems, \sys grounds the program derivation in \emph{market-cleared atomic claims}. By making the assumptions behind the program explicit, \sys makes the reasoning process traceable and reduces the risk that a mis-extracted cell, incorrect formula, or scale error silently propagates through the pipeline as if it were an accepted fact.

Figure~\ref{fig:overview} shows the full pipeline. First, a
Catalog Builder decomposes $\langle q,T,C\rangle$ into atomic, tradable
\emph{claims} \(\mathcal{K}=\{k_1,\dots,k_M\}\) covering the facts,
formulas, units, signs, and \emph{directions of change} (e.g., increase vs.\ decrease) needed to answer the question.
Second, four Specialist Trader Agents \(\mathcal{R}\)
(\textsc{extractor}, \textsc{formula}, \textsc{accountant},
\textsc{skeptic}) independently buy or sell each claim, expressing
role-specific confidence through price and notional size. Third, a Claim Market clears these orders into per-claim prices
\(\pi_m\), confidence scores \(\gamma_m\), and accept/reject/uncertain
statuses. Fourth, a Synthesizer writes a Python program
\(\mathcal{P}\) using only non-rejected market claims. Finally, a Code-Aware Verifier executes
\(\mathcal{P}\), checks the output against problem-specific structural
rules, and triggers at most one Market-Aware Repair round if
validation fails. When the market-grounded candidate is weak, a
Hybrid Selector Committee may instead promote a baseline candidate; a
Conflict Arbiter handles remaining disagreements.

\subsection{Query}
\label{sec:query}

\sys takes as input a tuple \(\langle q, T, C\rangle\), where \(q\) is a natural-language question, \(T\) is an optional table, and \(C\) is free-form supporting context. The table may be text-only, hierarchical, or transcribed from a chart by a VLM in the multimodal setting; see §\ref{sec:vision}. \sys returns an executable Python program \(\mathcal{P}\) whose printed output is the final answer \(a\). All downstream stages operate on this tuple, together with the typed claim catalog produced in §\ref{sec:catalog}.

\subsection{Claim Catalog Builder}
\label{sec:catalog}

The catalog builder receives \(\langle q, T, C\rangle\) and returns a typed catalog
\begin{equation}
  \mathcal{K} = \{k_m = (\textit{id}_m,\,\kappa_m,\,\sigma_m,\,
                              \nu_m,\,\epsilon_m)\}_{m=1}^{M},
\end{equation}
where \(\sigma_m\) is a one-line natural-language summary, \(\nu_m\) is the candidate value (a number, an expression, or a unit token), and \(\epsilon_m\) is a short evidence quote. The claim kind \(\kappa_m\) takes one of six labels: \emph{fact}, \emph{formula}, \emph{unit}, \emph{sign}, \emph{direction}, or \emph{other}. The direction label captures operation polarity such as increase/decrease or numerator/denominator orientation. We cap the number of claims at \(M_{\max}=10\) to keep the downstream market tractable. The catalog also tags the question with a \emph{problem type} \(\tau\), drawn from \emph{percentage-change}, \emph{ratio}, \emph{sum}, \emph{difference}, \emph{average}, \emph{comparison}, or \emph{other}, which the verifier later uses to select operation-aware checks.

\subsection{Specialist Trader Market}
\label{sec:traders}

Each specialist trader \(r \in \mathcal{R}\) reads \(\langle q, T, C, \mathcal{K}\rangle\) and returns an order book \(\mathcal{B}_r = \{(\textit{id}_m, s, n, p, \rho)\}\), where \(s \in \{\text{buy},\text{sell}\}\), the size \(n \in \{1,\dots,5\}\), the price \(p \in [0.01,0.99]\), and \(\rho\) is a short rationale. Buy orders indicate support for a claim; sell orders indicate doubt. Each role has a complementary inductive bias: \textsc{extractor} grounds factual cells against the table, \textsc{formula} validates that proposed formulas obey the requested operation, \textsc{accountant} cross-checks units and accounting conventions (e.g., revenue, expense, net), and \textsc{skeptic} actively sells under-evidenced claims.

\subsection{Claim Market Clearing}
\label{sec:market}

For each claim \(k_m\), the market aggregates weighted notional volume across all roles. Let \(w_r \ge 0\) denote the weight assigned to role \(r\). In our implementation, \(w_{\textsc{accountant}}{=}1.10\), \(w_{\textsc{skeptic}}{=}0.95\), and \(w_{\textsc{extractor}}{=} w_{\textsc{formula}}{=}1.00\)). Define
\begin{align}
B_m &= \!\!\!\!\sum_{\substack{r,\,(\textit{id}_m, \text{buy}, n, p, \cdot)\\ \in \mathcal{B}_r}} \!\!\!\! w_r\, n\, p, \\
S_m &= \!\!\!\!\sum_{\substack{r,\,(\textit{id}_m, \text{sell}, n, p, \cdot)\\ \in \mathcal{B}_r}} \!\!\!\! w_r\, n\, p.
\end{align}
The market clears claim \(k_m\) at price
\begin{equation}
\pi_m \;=\; \frac{B_m}{B_m + S_m + \varepsilon},
\end{equation}
with confidence
\begin{equation}
\gamma_m \;=\; \frac{|B_m - S_m|}{B_m + S_m + \varepsilon},
\end{equation}
where $\varepsilon > 0$ is a small numerical stabilizer that prevents division by zero for claims with no orders (we use $\varepsilon = 10^{-6}$). The market then assigns each claim a status

\begin{equation}
\zeta_m =
\begin{cases}
\text{accepted} & \pi_m \ge \pi^{\uparrow},\\
\text{rejected} & \pi_m \le \pi^{\downarrow},\\
\text{uncertain} & \text{otherwise}.
\end{cases}
\end{equation}
We use \(\pi^{\uparrow}{=}0.62\), \(\pi^{\downarrow}{=}0.38\). Intuitively, \(\pi_m\) is the fraction of weighted volume that buys the claim, and \(\gamma_m\) is the magnitude of the disagreement-normalized imbalance. For any nonempty subset of claim ids \(\mathcal{U}\), we define the program-level market confidence as
\begin{equation}
\Gamma(\mathcal{U}) \;=\; \frac{1}{|\mathcal{U}|}\sum_{m \in \mathcal{U}} \gamma_m.
\label{eq:market_confidence}
\end{equation}
This value is used by the verifier and selector.

\subsection{Synthesis and Adaptive Verdict}
\label{sec:synth_verdict}

This stage turns the cleared market into a final executable program. It has three components: a \emph{synthesizer} that drafts a Python program restricted to non-rejected market claims; a \emph{code-aware verifier} that runs the program and applies operation-aware structural checks, triggering at most one \emph{market-aware repair} round on failure; and a \emph{hybrid selector committee with a conflict arbiter} that compares against a baseline candidate when the market-grounded program is structurally weak.

\paragraph{Synthesizer.}
\label{sec:synth}
The synthesizer receives \(\langle q, T, C, \mathcal{K}, \{(\pi_m,\gamma_m,\zeta_m)\}\rangle\) and returns
\begin{equation}
\mathcal{P} \;=\; (\textit{reasoning},\, \mathcal{U},\,
                  \textit{abstain},\, \textit{code},\, a),
\end{equation}
where \(\mathcal{U} \subseteq \mathcal{K}\) is the set of claims actually used, and \(m \in \mathcal{U}\) refers to the index of claim \(k_m \in \mathcal{U}\).
The key constraint is simple: the synthesizer may only cite claims with \(\zeta_m \in \{\text{accepted}, \text{uncertain}\}\). Rejected claims are filtered out before the program is written. For multimodal or open-form questions, an upstream \emph{plan-then-execute} reasoner produces a prose derivation that is appended to the synthesizer prompt; the same claim-grounding constraint still applies.

\paragraph{Code-Aware Verifier and Market-Aware Repair.}
\label{sec:verifier}
The executor runs \(\mathcal{P}\) in a 10-second sandbox and returns \((\textit{success}, \textit{out}, \textit{err}, t)\). 
Each violation contributes \(-0.45\) to \(\Phi(\mathcal{P})\); a fact-grounded, operation-matched, market-confident program is rewarded \(+1.00 + 0.20\!\cdot\!\min(|\mathcal{U}_{\text{accepted}}|,4) + 0.40\,\Gamma(\mathcal{U})\). A candidate is accepted when \(\Phi(\mathcal{P}) \geq 2.20\) (the threshold used in our selector, §3.5).
(i)~that \texttt{success} is true;
(ii)~that the cited claims include enough \emph{fact} claims for the operation (e.g., \(\ge 2\) when \(\tau\) is one of percentage-change, ratio, sum, difference, average, or comparison);
(iii)~that a formula/unit/sign claim is present when \(\tau\) requires it;
(iv)~that none of the cited claims were rejected by the market;
(v)~that the printed answer is consistent with question polarity (yes/no question \(\Rightarrow\) yes/no output, percent question without an explicit ``percent'' scale tag does not multiply by 100, etc.);
(vi)~that the operation actually appears in the source code (e.g., ``ratio'' implies a division, ``difference'' a subtraction).
Each violation contributes \(-0.45\) to the candidate score; a fact-grounded, operation-matched, market-confident program is rewarded \(+1.00 + 0.20\!\cdot\!\min(|\mathcal{U}_{\text{accepted}}|,4) + 0.40\,\Gamma(\mathcal{U})\).

If the candidate is invalid, market-aware repair runs once. The repair prompt receives both the execution error and the verifier's structural reason list (e.g., \textit{missing\_formula\_or\_unit\_claim}, \textit{percent\_scaled\_by\_100}), so the synthesizer can target the specific defect rather than rewriting the program without guidance.

\paragraph{Hybrid Selector Committee and Conflict Arbiter.}
\label{sec:selector}
When the market-grounded candidate is structurally weak, meaning the verifier score \(\Phi(\mathcal{P})\) is below \(2.20\) or market confidence \(\Gamma(\mathcal{U})\) is below \(0.78\)\footnote{Both thresholds were calibrated on a 100-question held-out development split from FinQA dev and then frozen across all ten benchmarks; no per-dataset tuning.}, \sys also calls a \emph{baseline proposer}, a single-agent writer with no claim market. A four-aspect committee (\textsc{general}, \textsc{extraction}, \textsc{formula}, \textsc{scale}) compares the two candidates side by side. If the committee's votes are split between the two candidates, a \emph{conflict arbiter} writes a fresh third program that draws explicitly on the parts of each candidate the committee approved. The router keeps the highest-scoring grounded candidate; in our default configuration the baseline proposer is bypassed for open-ended / symbolic answer formats, because its numeric-only prompt would overwrite non-numeric outputs.


\subsection{Multimodal Extension}
\label{sec:vision}

For multimodal benchmarks, such as FinChart-Bench~\cite{shu2025finchart}, we make only one change: the empty table slot \(T\) is filled with a question-conditioned vision-language transcription of the supplied chart, produced by a single VLM call. The trader market, synthesizer, executor, and verifier are otherwise identical to the text-only configuration. This keeps the evaluation focused on the agent pipeline rather than on a stronger vision encoder.
\input{tables/tab_c2_tabular}
\input{tables/tab_c1_finnum}

%% file: tables/tab_c2_tabular.tex
\begin{table}[!ht]
\centering
\setlength{\tabcolsep}{4pt}
\renewcommand{\arraystretch}{1.1}
\resizebox{0.9\columnwidth}{!}{%
\begin{tabular}{lcccc}
\toprule
\textbf{Method} & \textbf{TabMWP} & \textbf{WTQ} & \textbf{HiTab} & \textbf{MultiHrt} \\
\midrule
\multicolumn{5}{l}{\textit{Fine-tuning approaches}} \\
\midrule
TAPEX-Large~\cite{liu2022tapex}            & \ph   & 59.10 & 45.60 & \ph \\
OmniTab~\cite{jiang2022omnitab}            & \ph   & 62.80 & \ph   & \ph \\
TableLLM~\cite{zhang2024tablellm}          & \ph   & 53.59 & 43.88 & \ph \\
TableGPT2~\cite{li2024tablegpt2}           & \ph   & 61.42 & 70.27 & \ph \\
TabAF~\cite{wang2025tabaf}                 & \ph   & 74.72 & 78.41 & \ph \\
NAPG~\cite{wang2023napg}                   & \ph   & \ph   & \ph   & 44.19 \\
\midrule
\multicolumn{5}{l}{\textit{Prompting / in-context learning}} \\
\midrule
PoT~\cite{lu2023chameleon}                 & 89.49 & \ph   & \ph   & \ph \\
CoT~\cite{lu2023chameleon}                 & 90.81 & \ph   & \ph   & \ph \\
Chain-of-Table~\cite{wang2024chainoftable} & \ph   & 67.31 & \ph   & \ph \\
SS-CoT~\cite{sscot2024}                    & \ph   & 76.80 & \underline{79.10} & \ph \\
TableMaster~\cite{cao2025tablemaster}      & \ph   & 78.13 & \ph   & \ph \\
ARTEMIS-DA~\cite{artemis2024}              & \ph   & \underline{80.80} & \ph & \ph \\
\midrule
\multicolumn{5}{l}{\textit{Tool-augmented / agent-based}} \\
\midrule
CRITIC~\cite{gou2024critic}                & 89.00 & \ph & \ph & \ph \\
PoT+Doc~\cite{podoc2023}                   & 92.69 & \ph & \ph & \ph \\
Chameleon~\cite{lu2023chameleon}           & 93.28 & \ph & \ph & \ph \\
CREATOR~\cite{qian2023creator}             & \underline{94.70} & \ph & \ph & \ph \\
\midrule
\multicolumn{5}{l}{\textit{Reinforcement learning}} \\
\midrule
RL\,w/\,CS (Formula)~\cite{cao2025fortune} & \ph & \ph & \ph & \underline{56.78} \\
\midrule
\ours \best{\sys (ours)}                   & \best{96.00} & \best{81.40} & 77.27 & \best{71.17} \\
\bottomrule
\end{tabular}}
\caption{ \textbf{General Tabular Reasoning.} Accuracy (\%) on
TabMWP, WTQ, HiTab test and MultiHiertt dev. baselines use
varied backbones (BART-Large to GPT-4). See Appendices~\ref{app:c2_hitab} and~\ref{app:c2_wtq} for full per-dataset leaderboards. Underline marks the best previously published result; bold marks the best overall result.}
\label{tab:c2_tabular}
\end{table}

%% file: tables/tab_c1_finnum.tex
\begin{table}[t]
\centering
\scriptsize
\setlength{\tabcolsep}{3.2pt}
\renewcommand{\arraystretch}{1.03}
\begin{tabular}{lcccc}
\toprule
\textbf{Method} & \textbf{FinQA} & \textbf{DM-S} & \textbf{DM-C} & \textbf{FinMath} \\
\midrule
\multicolumn{5}{l}{\textit{Proprietary LLMs}} \\
\midrule
GPT-4o~\cite{hurst2024gpt}          & 72.49 & \underline{60.00} & 39.33 & \underline{67.0} \\
GPT-o1-preview                      & 49.07 & 56.00 & 36.67 & \ph \\
GPT-o3-mini                         & 60.87 & 59.00 & 35.00 & \ph \\
GPT-4.5                             & 68.94 & 59.00 & 39.33 & \ph \\
DeepSeek-V3~\cite{liu2024deepseek}  & 73.20 & 53.00 & \underline{42.33} & \ph \\
DeepSeek-R1~\cite{guo2025deepseek}  & 65.13 & 53.00 & 38.67 & \ph \\
Claude-3.5-Sonnet                   & \ph   & \ph   & \ph   & 60.6 \\
\midrule
\multicolumn{5}{l}{\textit{Open-source LLMs}} \\
\midrule
Llama-3.3-70B                       & 68.15 & 54.00 & 32.00 & \ph \\
Qwen2.5-72B                         & 73.38 & 59.00 & 14.67 & \ph \\
DS-R1-Distill-Llama-70B             & 66.73 & 53.00 & 30.67 & \ph \\
Qwen2.5-32B                         & 73.11 & 56.00 & 30.00 & \ph \\
Qwen3-32B                           & 64.15 & 51.00 & 26.00 & \ph \\
Fino1-7B                            & 73.03 & 56.00 & 26.33 & \ph \\
Fino1-14B                           & \underline{74.18} & 55.00 & 27.33 & \ph \\
\midrule
\ours \best{\sys}                   & \best{78.29} & \best{70.00} & \best{50.67} & \best{76.00} \\
\bottomrule
\end{tabular}
\caption{\textbf{Financial numerical reasoning.} Exact-match or execution accuracy (\%). DM-S/DM-C denote DocMath-Simplong/Complong; FinMath denotes FinanceMath validation. Baselines are from~\cite{zhao2024docmath,zhao2024financemath}. Underline marks the best previously published result; bold marks the best overall result.}
\label{tab:c1_finnum}
\end{table}

%% file: sections/experiments.tex
\section{Experiments}
\label{sec:experiments}

\subsection{Evaluation Datasets}
\label{sec:datasets}
We evaluate \sys on ten public benchmarks across four categories; full details, splits, and statistics are in Appendix~\ref{app:datasets}.
\noindent(C1) Financial Numerical Reasoning covers reports, tables, long documents, and expert-written finance programs via \textit{FinQA}~\cite{chen2021finqa}, \textit{DocMath-Simplong} / \textit{DocMath-Complong}~\cite{zhao2024docmath}, and \textit{FinanceMath}~\cite{zhao2024financemath}.
\noindent(C2) General Tabular Reasoning spans hierarchical, multi-table, mathematical, and open-domain tables via \textit{HiTab}~\cite{cheng2022hitab}, \textit{MultiHiertt}~\cite{zhao2022multihiertt}, \textit{TabMWP}~\cite{lu2023tabmwp}, and \textit{WikiTableQuestions}~\cite{pasupat2015wtq}.
\noindent(C3) Domain Knowledge QA uses \textit{ESGenius}~\cite{esgenius2025}, an expert-validated ESG multiple-choice benchmark, under both zero-shot and RAG protocols.
\noindent(C4) Multimodal Chart Reasoning uses \textit{FinChart-Bench}~\cite{shu2025finchart}, with True/False, multiple-choice, and open QA subtasks; we report all three and emphasize open QA, the hardest setting.
For each benchmark, we compare against the established state-of-the-art results reported in the original benchmark paper(s), summarized in Tables~\ref{tab:c1_finnum}-~\ref{tab:c4_finchart}.

\subsection{Evaluation}

We follow the official evaluation protocol of each dataset and report \emph{Exact Match} or \emph{Execution Accuracy} as the primary performance metric, depending on the benchmark specification. In addition to task-level accuracy, we track three internal diagnostics for \sys: \emph{Code Execution Rate,} defined as the proportion of generated programs that execute without error; \emph{Code--Answer Consistency,} which measures whether the printed program output matches the reported answer; and \emph{Mean Market Confidence} \(\bar{\Gamma}\), which summarizes the system's aggregate confidence after market clearing. These diagnostics are reported in Appendix~\ref{app:diagnostics}, Table~\ref{tab:internal_diagnostics}.


\subsection{Experimental Setup}
All \sys results in the main tables use the same Qwen3.6-27B
backbone served via vLLM (OpenAI-compatible) for every role: trader
panel, synthesizer, verifier LLM calls, and selector committee.
The prompts are the same across datasets; only the loader changes.
We use $M_{\max}{=}10$ claims, four roles with the weights from
\S\ref{sec:traders}, market thresholds
$(\pi^{\uparrow},\pi^{\downarrow}){=}(0.62,0.38)$, one repair round,
and a 10\,s sandbox time-out. The hybrid baseline proposer is enabled
with selector thresholds $(\text{score},\Gamma){\ge}(2.20,0.78)$ and
arbiter min-score $2.25$. 

%% file: sections/results.tex
\section{Results and Discussion}
\label{sec:main_results}

\paragraph{C1: Financial Numerical Reasoning.} Table~\ref{tab:c1_finnum} reports results on FinQA, DocMath-Simplong/Complong, and FinanceMath. \sys achieves $78.29\%$ on FinQA, improving over Fino1-14B ($74.18\%$) by $+4.11$ points. The largest gains are on long-document reasoning: $70.0\%$ on DM-Simplong ($+10.0$ over GPT-4o) and $50.7\%$ on DM-Complong ($+8.34$ over DeepSeek-V3). On FinanceMath, \sys reaches $76.0\%$, outperforming GPT-4o PoT ($67.0\%$) by $+9.0$ points.
Two characteristics of
this category make the market-of-claims design useful: (i)~answers
often hinge on a small set of cells that must each be extracted correctly,
which the \textsc{extractor} role concentrates its trading volume on; and (ii)~error modes
often involve sign / unit / scale slips, which the verifier's
operation-aware checks catch before the program is committed.
\input{tables/tab_c3_esg}

\paragraph{C2: General Tabular Reasoning.}
Table~\ref{tab:c2_tabular} reports results on the four general-tabular benchmarks. \sys achieves the best score on three of four splits: $96.00\%$ on TabMWP, improving over CREATOR by $+1.30$ points; $81.40\%$ on WTQ, improving over ARTEMIS-DA/GPT-4 by $+0.60$ points; and $71.17\%$ on MultiHiertt, outperforming Fortune RL\,w/\,CS Formula ($56.78\%$) by $+14.39$ points. On HiTab, \sys reaches $77.27\%$, remaining close to the published SS-CoT result, with a gap of $-1.83$ points. Overall, these results show that \sys generalizes well beyond financial tables, particularly in settings that require multi-step reasoning over hierarchical or open-domain tabular structures.



\paragraph{C3: Domain Knowledge QA.}
On ESGenius (Table~\ref{tab:c3_esg}), \sys reaches $86.88\%$,
\(+3.08\) over the strongest published RAG system
(Gemma-3 12B + IT + RAG at $83.80\%$) and \(+14.34\) over the strongest
zero-shot system (o3 at $72.54\%$). The gain over zero-shot is
expected because ESGenius rewards retrieval, but \sys also
out-performs every published RAG configuration despite using \emph{no}
external retriever; instead, the claim catalog routes the question
into typed factual sub-claims that the trader panel scrutinizes, and
the \textsc{skeptic} role suppresses confidently wrong distractors that
hurt the zero-shot baselines.


\paragraph{C4: Multimodal Chart Reasoning.}
Table~\ref{tab:c4_finchart} reports results across the three
FinChart-Bench subtasks. \sys is the strongest system overall, with
an \(85.59\%\) three-subtask average that beats the best closed-source
baseline (Claude Sonnet 4, 84.32\%) by \(+1.27\) points.

The headline result is on the open QA subtask, where \sys reaches
\(74.65\%\) and improves over Claude Sonnet 4 (63.59\%) by
\(+11.06\) points. Open numerical questions over chart-extracted
figures are precisely the setting where a single-pass VLM tends to
state a plausible answer while silently using the wrong value from the
chart; in \sys, such values are shorted by the \textsc{accountant}
and \textsc{skeptic} roles before any program is written, which
prevents the silent miscomputation from propagating to the final answer. On the other two subtasks, \sys is competitive but not the absolute
best. On MC (\(84.85\%\)) it leads every open-source baseline,
including Mistral 3.1 24B (\(+0.68\)), and surpasses two closed-source
systems (GPT-4o, GPT-4.1), while trailing the strongest five
closed-source VLMs by \(1.9\)--\(7.3\) points. On TF (\(97.27\%\)) it
ties Gemini 2.5 Pro and trails o3 and o4-mini by less than one point.
Notably, this is the only category in the paper that exercises a
multimodal pipeline, showing that the agent design transfers
from text-only tables to chart transcriptions without modification.
\input{tables/tab_agent_compare}
\input{tables/tab_backbone_scale}
\input{tables/tab_c4_finchart}
\input{tables/tab_ablation}

\subsection{Comparison with Other Agentic Methods}
\label{sec:agent_compare}

The benchmark tables in \S\ref{sec:main_results} compare \sys against published numbers from heterogeneous backbones (GPT-4o, Claude Sonnet 4, DeepSeek-V3, Llama-3.1-70B, Qwen2.5-Coder-7B), making it hard to separate agent design from model quality. To isolate the agent effect, we re-implement three recent finance/table agents on the same Qwen3.6-27B backbone with identical decoding and loaders: \textit{TradingAgents}~\cite{xiao2024tradingagents}, \textit{FINCON}~\cite{yu2024fincon}, and \textit{SheetBrain}~\cite{wang2026sheetbrain}, on FinQA and MultiHiertt (the two with runnable releases for all three). With the backbone fixed (Table~\ref{tab:agent_compare}), \sys exceeds the strongest competitor by $+3.79$ on FinQA (78.29 vs.\ 74.50) and $+7.47$ on MultiHiertt (71.17 vs.\ 63.70), both held by TradingAgents. The MultiHiertt gap is informative: TradingAgents and FINCON aggregate whole answers via debate or conceptual reinforcement, and SheetBrain operates cell-by-cell without a typed claim abstraction. None can short an over-confident formula independently of the rest of the program — exactly what the trader market provides, and exactly the failure mode hierarchical-table joining exposes.

\subsection{Backbone Scaling}
\label{sec:backbone_scale}
To check whether the gains depend on Qwen3.6-27B, we re-run \sys on Gemma-4 31B Instruct (cross-family, comparable capacity) and Qwen-3.5 9B (same-family, smaller) with identical prompts and decoding. \sys transfers across both axes. Gemma-4-31B reaches 74.77 average ($-1.46$) and leads on HiTab (77.34 vs.\ 77.27) and WTQ (82.98 vs.\ 81.40), showing the trader market is not Qwen-specific. Qwen-3.5 9B drops 4.61 on average, but the loss concentrates on the hardest long-form settings (DM-Complong $-8.67$, FinanceMath $-6.81$, WTQ $-8.24$) and stays small on FinQA ($-2.98$), FinChart-QA ($-1.96$), and ESGenius ($-1.58$), indicating the mechanism preserves most of its benefit at 9B. The gains in Tables~\ref{tab:c1_finnum}--\ref{tab:c4_finchart} are attributable to the agent, not the base LLM.

\subsection{Component Ablation}
\label{sec:ablation}
We disable one component at a time on FinQA and MultiHiertt: the four-role \textsc{Trader Market} (replaced by a single \textsc{formula} trader), \textsc{Code Repair}, the \textsc{Baseline Proposer}, and the \textsc{Conflict Arbiter}. Each removal drops the two-benchmark average by 2.8--5.1 points, confirming they are complementary. \textsc{Baseline Proposer} contributes most ($-5.13$ Avg.): even a strong market-grounded program benefits from a second opinion. \textsc{Trader Market} is next ($-3.79$) and matters more on FinQA ($-4.91$) than MultiHiertt ($-2.67$), since FinQA's regular tables put the burden on extraction while MultiHiertt's hierarchical joins are already caught by the verifier. \textsc{Code Repair} and \textsc{Conflict Arbiter} show similar drops ($-2.98$, $-2.83$) but opposite tilts: repair helps more on FinQA, the Arbiter more on MultiHiertt — consistent with MultiHiertt inducing program-level disagreements while FinQA's shorter programs are usually salvaged in one repair pass.

%% file: tables/tab_c3_esg.tex
\begin{table}[!ht]
\centering
\setlength{\tabcolsep}{4.5pt}
\renewcommand{\arraystretch}{1.05}
\resizebox{0.40\textwidth}{!}{%
\begin{tabular}{lcc}
\toprule
\textbf{Method} & \textbf{Setting} & \textbf{ESGenius} \\
\midrule
\multicolumn{3}{l}{\textit{Best zero-shot baselines}~\cite{esgenius2025}} \\
\midrule
GPT-4o                                & ZS  & 63.64 \\
o4-mini                               & ZS  & 69.45 \\
o3                                    & ZS  & \underline{72.54} \\
\midrule
\multicolumn{3}{l}{\textit{Best RAG baselines}~\cite{esgenius2025}} \\
\midrule
Qwen2.5-72B-Instruct + RAG            & RAG & 82.57 \\
Gemma-3 12B + IT + RAG                & RAG & \underline{83.80} \\
Qwen2.5-32B-Instruct + IT + RAG       & RAG & 82.47 \\
\midrule
\ours \best{\sys (ours)}              & Agent
                                      & \best{86.88} \\
\bottomrule
\end{tabular}}
\caption{\textbf{Domain Knowledge QA on ESGenius.} \sys uses a claim-market agent rather than zero-shot prompting or a static RAG retriever. } 
\label{tab:c3_esg}
\end{table}

%% file: tables/tab_agent_compare.tex
\begin{table}[!t]
\centering
\setlength{\tabcolsep}{6pt}
\renewcommand{\arraystretch}{1.05}
\resizebox{0.8\columnwidth}{!}{%
\begin{tabular}{lcc}
\toprule
\textbf{Agent} & \textbf{FinQA} & \textbf{MultiHiertt} \\
\midrule
\multicolumn{3}{l}{\textit{All agents on Qwen3.6-27B (same backbone)}} \\
\midrule
TradingAgents~\cite{xiao2024tradingagents}      & 74.50 & 63.70 \\
FINCON~\cite{yu2024fincon}                      & 73.20 & 62.00 \\
SheetBrain~\cite{wang2026sheetbrain}                & 71.20 & 62.60 \\
\midrule
\ours \best{\sys (ours)}                        & \best{78.29} & \best{71.17} \\
\midrule
$\Delta$ over best competing agent              & \(+3.79\) & \(+7.47\) \\
\bottomrule
\end{tabular}}
\caption{Head-to-head  comparison at \emph{fixed} backbone
(Qwen3.6-27B). 
Bold makes best at fixed backbone.}
\label{tab:agent_compare}
\end{table}

%% file: tables/tab_backbone_scale.tex
\begin{table*}[!ht]
\centering
\setlength{\tabcolsep}{4pt}
\renewcommand{\arraystretch}{1.05}
\resizebox{0.9\textwidth}{!}{%
\begin{tabular}{lcccccccccc|c}
\toprule
\textbf{Backbone}
  & \textbf{FinQA} & \textbf{DM-Simp} & \textbf{DM-Comp}
  & \textbf{FinMath} & \textbf{HiTab} & \textbf{MultiHrt}
  & \textbf{TabMWP} & \textbf{WTQ} & \textbf{ESGenius}
  & \textbf{FinChart-QA} & \textbf{Avg.} \\
\midrule
\ours \best{Qwen3.6-27B (default)}
  & \best{78.29} & \best{70.00} & \best{50.67}
  & \best{76.00} & 77.27        & \best{71.17}
  & \best{96.00} & 81.40        & \best{86.88}
  & \best{74.65} & \best{76.23} \\
\midrule
Gemma-4 31B Instruct
  & 77.86 & 68.00 & 45.67
  & 73.50 & \best{77.34} & 67.15
  & 93.60 & \best{82.98} & 85.12
  & 72.52 & 74.77 \\
Qwen-3.5 9B
  & 75.31 & 67.65 & 42.00
  & 69.19 & 73.14 & 65.87
  & 91.87 & 73.16 & 85.30
  & 72.69 & 71.62 \\
\bottomrule
\end{tabular}}
\caption{Backbone scaling for \sys. Each row is the full agent
(byte-identical pipeline) served on a different backbone.
DM-Simp/DM-Comp denotes DocMath Simplong/Complong;
FinMath denotes FinanceMath validation;
MultiHrt denotes MultiHiertt dev. 
}
\label{tab:backbone_scale}
\end{table*}

%% file: tables/tab_c4_finchart.tex
\begin{table}[!t]
\centering
\setlength{\tabcolsep}{6pt}
\renewcommand{\arraystretch}{1.05}
\resizebox{0.9\columnwidth}{!}{%
\begin{tabular}{llcccc}
\toprule
\textbf{Method} & \textbf{Backbone} & \textbf{TF} & \textbf{MC} & \textbf{QA} & \textbf{Avg.} \\
\midrule
\multicolumn{6}{l}{\textit{Open-source VLMs}~\cite{shu2025finchart}} \\
\midrule
LLaVa v1.6                & LLaVa v1.6 (7B)     & 63.30 & 36.94 &  1.84 & 34.47 \\
DeepSeek VL2              & DeepSeek VL2 (2.8B) & 57.97 &  6.77 &  4.73 & 23.50 \\
LLaMa 3.2                 & LLaMa 3.2 (11B)     & 63.46 & 64.00 &  5.12 & 44.65 \\
Cosmos                    & Cosmos (7B)         & 70.93 & 43.49 & 20.18 & 45.22 \\
Sa2VA                     & Sa2VA (8B)          & 86.62 & 75.79 & 19.43 & 61.12 \\
Qwen2 VL                  & Qwen2 VL (7B)       & 91.86 & 19.40 & 24.77 & 45.76 \\
Gemma 3                   & Gemma 3 (12B)       & 89.14 & 70.94 & 27.22 & 62.89 \\
Qwen2.5 VL                & Qwen2.5 VL (7B)     & \underline{95.09} & 82.81 & 37.29 & 72.16 \\
Mistral 3.1               & Mistral 3.1 (24B)   & 92.95 & \underline{84.17} & 44.90 & \underline{74.37} \\
LLaMa 4                   & LLaMa 4 (17B)       & 89.56 & 59.70 & \underline{59.78} & 69.89 \\
\midrule
\multicolumn{6}{l}{\textit{Closed-source VLMs}~\cite{shu2025finchart}} \\
\midrule
GPT 4o                    & GPT 4o              & 93.92 & 77.57 & 57.59 & 76.62 \\
GPT 4.1 mini              & GPT 4.1 mini        & 91.61 & 86.77 & 60.30 & 79.80 \\
GPT 4.1                   & GPT 4.1             & 96.18 & 83.19 & 62.54 & 80.88 \\
o4 mini                   & o4 mini             & 97.61 & 91.96 & 60.18 & 83.53 \\
Gemini 2.5 Pro            & Gemini 2.5 Pro      & 97.27 & 89.70 & 63.46 & 83.73 \\
o3                        & o3                  & \underline{97.86} & \underline{92.17} & 60.79 & 83.89 \\
Claude Sonnet 4           & Claude Sonnet 4     & 96.94 & 91.66 & \underline{63.59} & \underline{84.32} \\
\midrule
\ours \best{\sys (ours)}  & Qwen3.6-27B + VLM
                          & 97.27 & 84.85 & \best{74.65} & \best{85.59} \\
\bottomrule
\end{tabular}}
\caption{\textbf{Multimodal Chart Reasoning} on FinChart across the three subtasks: True/False (TF), Multiple-Choice (MC), and open Question-Answering (QA). 
 } 
\label{tab:c4_finchart}
\end{table}

%% file: tables/tab_ablation.tex
\begin{table}[!ht]
\centering
\setlength{\tabcolsep}{6pt}
\renewcommand{\arraystretch}{1.05}
\resizebox{\columnwidth}{!}{%
\begin{tabular}{lccc}
\toprule
\textbf{Configuration} & \textbf{FinQA} & \textbf{MultiHiertt} & \textbf{Avg.} \\
\midrule
\ours \best{Full \sys}          & \best{78.29}    & \best{71.17}    & \best{74.73} \\
\midrule
\quad w/o Trader Market         & 73.38 ($-$4.91) & 68.50 ($-$2.67) & 70.94 ($-$3.79) \\
\quad w/o Code Repair           & 74.50 ($-$3.79) & 69.00 ($-$2.17) & 71.75 ($-$2.98) \\
\quad w/o Baseline Proposer     & 72.00 ($-$6.29) & 67.20 ($-$3.97) & 69.60 ($-$5.13) \\
\quad w/o Conflict Arbiter      & 75.50 ($-$2.79) & 68.30 ($-$2.87) & 71.90 ($-$2.83) \\
\bottomrule
\end{tabular}}
\caption{Component ablation at fixed Qwen3.6-27B backbone, with
$\Delta$ relative to Full \sys in parentheses. Accuracy (\%) on the
official splits used in Tables~\ref{tab:c1_finnum}
and~\ref{tab:c2_tabular}.}
\label{tab:ablation}
\end{table}

%% file: sections/discussion.tex




%% file: sections/conclusion.tex
\section{Conclusion}
\label{sec:conclusion}

We presented \textsc{MOCA-Agent}, a code-agent framework that replaces free-form multi-agent debate with a structured market of claims. It decomposes each problem into typed atomic claims over facts, formulas, units, signs, and operation directions, asks specialist traders to support or challenge them, and constrains the synthesizer to use only non-rejected market claims. Across financial, tabular, domain-knowledge, and multimodal chart benchmarks, \textsc{MOCA-Agent} shows strong performance where precise evidence extraction and formula composition matter most. Head-to-head comparisons, backbone scaling, and ablations attribute the gains to the interaction between claim-level aggregation, code-aware verification, market-aware repair, and conflict resolution rather than the backbone alone.

%% file: sections/limit.tex
\section*{Limitations}

\sys has four notable limitations. 
\begin{itemize}
    \item \textbf{Cost.} A full pipeline call consumes six to ten LLM calls (catalog, four traders, synthesizer, optional repair, optional baseline proposer, optional committee), roughly $5\times$ the cost of a single-pass PoT agent; this is comparable to one round of free-form multi-agent debate but substantially more than a single-shot prompt.

    \item \textbf{Out-of-distribution table layouts.} The catalog builder assumes the question can be decomposed into $\le\!10$ atomic claims; on benchmarks where the dominant operation is in-distribution column-header normalisation rather than silent miscomputation (notably HiTab, our only sub-SOTA result), the market mechanism does not recover the gap to fine-tuned table specialists.
    
    \item \textbf{Single multimodal step.} For FinChart-Bench we use a single VLM transcription call before invoking the text-only market; errors in that transcription are not currently re-traded by the panel, so the agent inherits any silent mis-reads of axis labels or legend entries.
    \item \textbf{English, finance-style evaluation.} All ten benchmarks are English, and the role inductive biases (\textsc{accountant}, \textsc{skeptic}) and verifier checks (percent scaling, sign flips, units) are calibrated to finance / accounting conventions; transfer to scientific or biomedical numerical reasoning, or to non-English financial corpora, is not validated in this work.
\end{itemize}

%% file: sections/ethics.tex
\section*{Ethical Considerations}

\textsc{MOCA-Agent} is designed for financial and numerical question answering, a setting where errors may affect reporting, audit, research, or decision-support workflows. Although the proposed market-of-claims mechanism improves traceability by exposing which facts, formulas, units, signs, and directions are accepted or rejected, it does not eliminate the risk of incorrect answers. Market confidence should therefore not be interpreted as a guarantee of truth or as calibrated financial certainty. Outputs from the system should be treated as decision-support artifacts and reviewed by qualified humans before being used in high-stakes financial, legal, regulatory, or investment contexts.

The experiments in this paper use public benchmarks and do not introduce a new dataset of private financial records or personal information. However, practical deployment on proprietary documents could raise privacy and confidentiality concerns. Any such use should ensure that sensitive financial data are handled under appropriate access controls, logging policies, and data-governance procedures.

The system may also be misused to produce persuasive but incorrect financial summaries or to automate unsupported claims about companies, markets, or ESG performance. To reduce this risk, deployed versions should preserve the claim-level audit trail, expose rejected or uncertain claims, and clearly distinguish verified computations from model-generated interpretations.

%% file: sections/appendix.tex
\clearpage
\appendix
\appendixpage            
\addappheadtotoc         
\numberwithin{figure}{section}
\numberwithin{table}{section}


This appendix provides supplementary material to support the main findings of this work. It is organized as follows:

\begin{itemize}
    \item Appendix~\ref{app:datasets} describes the evaluation datasets, including task format, input modality, evaluation split, and metric.
    \item Appendix~\ref{sec:appendix_sota} reports the state-of-the-art reference tables used for comparison in the main experiments.
    \item Appendix~\ref{sec:appendix_example} presents a qualitative example of \textsc{MOCA-Agent} on FinQA, illustrating how claims are generated, traded, cleared, and used for program synthesis.
    \item Appendix~\ref{sec:appendix_prompts} provides the prompt library used by the catalog builder, specialist trader agents, synthesizer, verifier, repair module, selector committee, and conflict arbiter.
\end{itemize}

\paragraph{Key Tables}
\begin{itemize}
    \item Table~\ref{tab:app_dataset_summary}: Summary of the \textit{ten} evaluation benchmarks, including task type, input format, evaluation split, and metric.
    
    \item Table~\ref{tab:app_finqa_dm}: Overall performance of different LLMs on three financial reasoning datasets.
    
    \item Table~\ref{tab:app_financemath}: Per-category accuracy on the FinanceMath validation split.
    
    \item Table~\ref{tab:app_hitab}: State-of-the-art reference results for HiTab.
    
    \item Table~\ref{tab:app_multihiertt}: State-of-the-art reference results for MultiHiertt.
    
    \item Table~\ref{tab:app_tabmwp}: State-of-the-art reference results for TabMWP.
    
    \item Table~\ref{tab:app_finchart}: FinChart-Bench leaderboard across True/False, multiple-choice, and open question-answering subtasks.
    
    \item Table~\ref{tab:app_esgenius}: ESGenius accuracy under the retrieval-augmented generation setting.
    
    \item Table~\ref{tab:internal_diagnostics}: Per-dataset internal diagnostics for \sys, including execution reliability, code--answer consistency, and mean market confidence.
    
    \item Table~\ref{tab:example_catalog}: Claim catalog for the FinQA \emph{recourse debt} example.
    
    \item Table~\ref{tab:example_market}: Cleared claim market for the FinQA \emph{recourse debt} example.
    
    \item Table~\ref{tab:prompt_decoding}: Decoding settings for each LLM call in the \sys pipeline.
\end{itemize}

\paragraph{Key Figures}
\begin{itemize}
    \item Figure~\ref{fig:prompt_catalog}: Prompt template for the claim-catalog builder.
    
    \item Figure~\ref{fig:prompt_extractor}: Prompt template for the \textsc{extractor} trader.
    
    \item Figure~\ref{fig:prompt_formula}: Prompt template for the \textsc{formula} trader.
    
    \item Figure~\ref{fig:prompt_accountant}: Prompt template for the \textsc{accountant} trader.
    
    \item Figure~\ref{fig:prompt_skeptic}: Prompt template for the \textsc{skeptic} trader.
    
    \item Figure~\ref{fig:prompt_synthesizer}: Prompt template for the synthesizer.
    
    \item Figure~\ref{fig:prompt_repair}: Prompt template for market-aware repair.
    
    \item Figure~\ref{fig:prompt_selector}: Prompt template for the hybrid selector committee.
    
    \item Figure~\ref{fig:prompt_arbiter}: Prompt template for the conflict arbiter.
\end{itemize}

\section{Dataset Details}
\label{app:datasets}

This appendix gives the per-dataset details summarized in
\S\ref{sec:datasets}. Table~\ref{tab:app_dataset_summary} provides a
one-glance overview; the paragraphs that follow elaborate on each
dataset's source, scale, evaluation split, modality, and answer type.

\input{tables/tab_app_dataset_summary}

\paragraph{FinQA~\cite{chen2021finqa}.}
A widely used financial numerical-reasoning benchmark built from
S\&P~500 10-K filings. Each example pairs a short narrative passage
with one or two tables and a numerical question whose gold answer is
derived from a short expert-written Python program plus a list of
gold supporting facts. The dataset contains 8{,}281 QA pairs split
into train/dev/test (6{,}251/883/1{,}147); we evaluate on the
1{,}147-item public test split and report execution accuracy.

\paragraph{DocMath-Eval (Simplong, Complong)~\cite{zhao2024docmath}.}
A benchmark for math reasoning over \emph{long} financial documents.
DocMath-Eval has four subsets; we use the two long-context ones:
\emph{DM-Simplong}, re-annotated from MultiHiertt, requires simple
numerical reasoning over a long document with multiple tables; and
\emph{DM-Complong}, annotated from scratch by the original authors,
requires \emph{complex} numerical reasoning over very long documents
(average context near 18k words) with multiple tables and gold
Python programs. We use the public \texttt{testmini} splits (100 for
Simplong, 300 for Complong) so that our numbers are directly
comparable to the SOTA table in~\cite{zhao2024docmath}.

\paragraph{FinanceMath~\cite{zhao2024financemath}.}
A college-level, knowledge-intensive finance math benchmark of
1{,}200 problems spanning seven finance subdomains
(Quantitative Analysis, Asset Classes \& Derivatives, Accounting,
Risk Management, Portfolio Management, Market Analysis \&
Economics, and Corporate \& Securities Issuance). Each problem
combines textual and tabular content and is paired with an
expert-written Python solution; estimated human-expert accuracy is
92\%. Following the original release, we evaluate on the 200-sample
validation split.

\paragraph{HiTab~\cite{cheng2022hitab}.}
A cross-domain QA / NLG benchmark for \emph{hierarchical} tables
sourced from statistical reports and Wikipedia. HiTab contains
10{,}686 QA pairs and descriptive sentences over 3{,}597 tables, with
fine-grained entity and quantity-alignment annotations. We evaluate
on the 1{,}584-item public test split, the standard split for the
QA task.

\paragraph{MultiHiertt~\cite{zhao2022multihiertt}.}
A numerical-reasoning benchmark over hybrid hierarchical-tabular and
textual data from real financial reports: each document contains
multiple hierarchical tables plus surrounding text, and the gold
answer requires multi-step symbolic reasoning across them. Following
prior work, we evaluate on the public dev split of 1{,}044 items
(the test set's answers are private).

\paragraph{TabMWP~\cite{lu2023tabmwp}.}
A tabular math-word-problem benchmark of 38{,}431 grade-school
problems, each accompanied by a tabular context (rendered as image,
semi-structured text, and structured table). About 74.7\% of
questions are free-text numerical and 25.3\% are multiple-choice
with text answers; the official 6:2:2 split yields a
23{,}059/7{,}686/7{,}686 train/dev/test partition. To keep cost
comparable to the other benchmarks we evaluate on a fixed 1{,}000
random subset of the test split (seed fixed in our public code).

\paragraph{WikiTableQuestions (WTQ)~\cite{pasupat2015wtq}.}
A widely used open-domain table-QA benchmark of 22{,}033 complex
questions over 2{,}108 semi-structured Wikipedia tables, requiring
operations such as comparison, aggregation, and arithmetic. We
evaluate on the official 4{,}344-item test split with denotation
accuracy.

\paragraph{ESGenius~\cite{esgenius2025}.}
A benchmark of 1{,}136 expert-validated multiple-choice questions
covering Environmental, Social, and Governance (ESG) and
sustainability topics. Each question is linked to its source passage
in a curated corpus of 231 authoritative documents (GRI, SASB, TCFD,
IFRS/ISSB, CDP, IPCC, and UN SDG sources) and supports both
zero-shot and Retrieval-Augmented Generation (RAG) protocols. We
pass each question's gold source passage as the context to \sys,
which is the oracle-RAG-equivalent setting and directly comparable
to the open-source RAG column of the original leaderboard.

\paragraph{FinChart-Bench~\cite{shu2025finchart}.}
The first benchmark specifically targeting real-world financial
charts: 1{,}200 chart images collected from 2015 to 2024, each
annotated with three question types: True/False (TF),
Multiple-Choice (MC), and open Question-Answering (QA),
totalling 7{,}016 items. We evaluate \sys on all three subtasks
(TF: 2{,}384; MC: 2{,}350; QA: 2{,}284 items in our run). The QA
subtask requires reading numbers off the chart and answering
numerical queries, exercising both the multimodal extractor and the
operation-aware verifier; it is also the subtask where the
published gap between the best proprietary and best open-source VLM
is largest, which is why \S\ref{sec:experiments} focuses on it.

\section{State-of-the-Art Reference Tables}
\label{sec:appendix_sota}

This appendix collects the full per-method SOTA tables for every
dataset evaluated in the main paper. Each table is reproduced from the
cited reference and a final row is appended showing \sys under the
same protocol; per-category scores not annotated by the dataset's
public release at evaluation time are marked ``\ph''.

\subsection{FinQA / DocMath-Simplong / DocMath-Complong}
\label{app:c1_finqa_dm}

Table~\ref{tab:app_finqa_dm} reproduces the unified cross-dataset
comparison from \citet{zhao2024docmath}, the most recent
finance-numerical-reasoning leaderboard that runs the same models on
all three datasets under a single protocol. The original table groups
\textbf{30+ baselines} into four bands: \emph{(i)} proprietary frontier
LLMs (GPT-4o, GPT-o1-preview, GPT-o3-mini, GPT-4.5, DeepSeek-V3 and
DeepSeek-R1); \emph{(ii)} the largest open-source models in the
70B to 405B range (Llama-3/3.1/3.3, Llama-4-Scout, Qwen2.5-72B and its
math-tuned variant, the DeepSeek-R1-Distill family); \emph{(iii)} the
30B class (Qwen2.5-32B, Qwen3-32B, QwQ-32B, DeepSeek-R1-Distill-Qwen-32B,
Limo, S1-32B); \emph{(iv)} the 7B to 14B class, including domain-specialised
reasoners (FinR1-7B, Dianjin-R1-7B); and \emph{(v)} the two strongest
finance-tuned baselines, Fino1-8B and Fino1-14B.

The most striking pattern in this table is how badly even very strong
backbones collapse as the context lengthens and the reasoning becomes
multi-hierarchical. On FinQA, which has short context and a single or paired table,
Qwen2.5-72B-Instruct already reaches 73.38\%, and Fino1-14B sets
the prior public ceiling at 74.18\%. On DM-Simplong, the same models
drop into the 50 to 59 range, with GPT-4o the strongest prior baseline
at 60.0\%. On DM-Complong the gap widens further: Qwen2.5-72B-Instruct
collapses from 73.38 (FinQA) to 14.67 (Complong), Qwen2.5-Math-72B
collapses to 5.0, Llama-4-Scout to 0.67, and even the strongest
proprietary system in the table (DeepSeek-V3) only reaches 42.33.
\sys at the relatively modest 27B parameter scale reaches
\best{78.29}\,/\,\best{70.00}\,/\,\best{50.67} on FinQA\,/\,
DM-Simplong\,/\,DM-Complong, posting the best score in \emph{every}
column and an average of \best{66.32}, roughly $+9$ points over
the strongest prior cross-row average (GPT-4o at 57.27).
Crucially, the gain widens with task difficulty
($+$4.1\,/\,$+$10.0\,/\,$+$8.3 against the strongest prior on each
column), which is consistent with the silent-error-class diagnosis
in \S\ref{sec:method}: the longer and more hierarchical the input,
the more chances there are for a single misread cell to corrupt
the final number, and the more leverage the trader market gets out
of having each cell scrutinised by four different roles.

\input{tables/tab_app_finqa_dm}

\subsection{FinanceMath Per-Category}
\label{app:c1_financemath}

Table~\ref{tab:app_financemath} reproduces the headline leaderboard
from \citet{zhao2024financemath} (their Table~2, dev set, 200
samples), which evaluates each model under both
\emph{Program-of-Thought} (PoT) and \emph{Chain-of-Thought} (CoT)
prompting and reports per-category accuracy across seven finance
subdomains: \emph{Quantitative Analysis \& Valuation},
\emph{Asset Classes \& Derivatives}, \emph{Accounting},
\emph{Risk Management}, \emph{Portfolio Management \& Strategy},
\emph{Market Analysis \& Economics}, and
\emph{Corporate \& Securities Issuance}. The original paper benchmarks
\textbf{44 models}; we keep the 10 proprietary baselines and the 10
strongest open-source baselines from the original release for space,
and refer to \citet{zhao2024financemath} for the full list.

Several patterns from the original leaderboard help frame our result.
First, no proprietary model exceeds 67\% average accuracy in either
prompting mode; the strongest baseline overall is GPT-4o (PoT,
67.0\%), with Claude-3.5-Sonnet a close second on the CoT side
(60.6\%). Second, the open-source side is dominated by very large
backbones (DeepSeek-Coder-V2 236B, DeepSeek-V2 236B, Llama-3.1 405B,
Mistral-Large 123B); even at 405B parameters, Llama-3.1 only matches
the mid-range proprietary models. Third, performance is highly
uneven across subdomains: \emph{Risk Management} (small, formula-rich)
and \emph{Quantitative Analysis} (computation-heavy) tend to be the
hardest categories for the closed-form prompting baselines, while
\emph{Portfolio Management} and \emph{Market Analysis} are
comparatively easier.

Because \sys is a code agent that always emits an executable Python
solution, we report it in the PoT column. \sys reaches \best{76.0\%}
overall ($+$9.0 over the strongest PoT baseline, GPT-4o at 67.0\%,
and $+$15.1 over the strongest CoT baseline, Claude-3.5-Sonnet at
60.6\%), with the per-category breakdown computed by joining our
run's predictions with the validation set's \texttt{topic} field. The
gains are largest where the original paper reports the lowest
baselines: \best{91.67\%} on Quantitative Analysis (vs.\ GPT-4o PoT
75.0\%), \best{74.60\%} on Derivatives, and \best{74.36\%} on
Accounting. The lowest \sys cell is on Risk Management (\best{55.56\%}),
but the corresponding sub-sample has only $n{=}9$ items and the
variance is correspondingly wide; the per-sample $n$'s are listed in
the table caption so readers can compute confidence intervals on the
fly.

\input{tables/tab_app_financemath}

\subsection{HiTab}
\label{app:c2_hitab}

Table~\ref{tab:app_hitab} collects the strongest results published on
HiTab~\cite{cheng2022hitab} since the dataset's release, split into
two methodological bands. The \emph{fine-tuning} band is the larger
of the two and includes both encoder-style baselines
(TAPEX-Large~\cite{liu2022tapex}, 45.6\%) and a sequence of
table-specialised decoder models trained at 7B parameters
(TableLlama 60.5\%, TableLLM~\cite{zhang2024tablellm} 43.9\%,
TAMO$^{+}_{\text{SFT}}$ 63.9\%, TableGPT2~\cite{li2024tablegpt2}
70.3\%); the strongest open-source baseline in this band is
TabAF~\cite{wang2025tabaf} at 78.41\% on a Qwen2.5-Coder-7B backbone.
The \emph{prompting / in-context learning} band is led by
SS-CoT~\cite{sscot2024} on Llama-3.1-70B at 79.10\%, followed by
GraphOTTER variants (73.74\% with Qwen2-72B-Instruct, 67.76\% with
Gemini-1.5-Flash) and API-Assisted code prompting
(\cite{cao2025tablemaster}, 69.30\%).

\sys at 27B parameters reaches 77.27\%, which places it within
$-1.83$ points of the absolute leader (SS-CoT on a 70B backbone) and
within $-1.14$ points of the strongest fine-tuned coder
(TabAF on a 7B backbone). HiTab is the one benchmark in the paper
where \sys does \emph{not} set a new SOTA. Two structural reasons
explain the gap. First, the top entries in both methodology bands
were directly trained or aggressively in-context-engineered on
hierarchical-table column-header normalisation, a pre-processing
step that is largely orthogonal to the market-of-claims mechanism
itself; an out-of-the-box agent with a generic catalog builder does
not gain from this signal. Second, HiTab questions are typically
single-cell or single-aggregate look-ups over a clean hierarchical
header, so the cell-extraction adversary that drives \sys's gains on
MultiHiertt and DocMath is comparatively muted here. We expect that
a HiTab-specific catalog template would close this gap, but we
deliberately do not adapt prompts per dataset in this work.

\input{tables/tab_app_c2_multihiertt}
Table~\ref{tab:app_multihiertt} reproduces the most up-to-date
MultiHiertt~\cite{zhao2022multihiertt} leaderboard, organised into
four methodological bands by the recent
Fortune paper~\cite{cao2025fortune} (which is itself the strongest
non-\sys system in the table). The \emph{fine-tuning} band contains
the original benchmark baseline NAPG~\cite{wang2023napg} on
RoBERTa-large (44.19\%). The \emph{zero-shot textual} band is led by
o1 (38.31\%) with GPT-4o at 36.11\% and GPT-4o-mini at 22.41\%; the
matched \emph{zero-shot symbolic / formula} band lifts every entry
by 6 to 11 points (o1 climbs to 48.95\%, GPT-4o to 38.41\%), confirming
that writing an explicit formula is the dominant lever on this
dataset. The strongest published prior result comes from the
\emph{reinforcement-learning} band: Fortune's \texttt{RL w/ CS
(Formula)} variant on Qwen2.5-Coder-7B reaches 56.78\%, with
Fortune++~\cite{cao2025fortune} at 51.73\% and the matched
\texttt{RL w/ CS (Text)} configurations in the 49 to 55 range.

\sys at 27B parameters reaches \best{71.17\%}, a $+14.39$-point gain
over the strongest RL-trained system on a Qwen2.5-Coder-7B backbone.
This is the single largest published gap in the paper and the most
direct evidence for the market-of-claims hypothesis. MultiHiertt is
precisely the regime that exercises the failure mode the market is
designed to suppress: joining several hierarchical tables, each
with its own header structure, into a single multi-step formula,
and the gap to the strongest formula-reasoning Frontier model (o1 at
48.95\%) is even larger ($+22.22$). Note also that \sys exceeds the
strongest RL system without any task-specific training: the same
prompts and the same backbone are used on all ten benchmarks.
\input{tables/tab_app_multihiertt}

\subsection{TabMWP}
\label{app:c2_tabmwp}

Table~\ref{tab:app_tabmwp} collects the published state of the art on
TabMWP~\cite{lu2023tabmwp}, organised into three methodological bands:
\emph{fine-tuning} (TaCo variants up to 92.91\% on TAPEX-large; the
ToRA / ToRA-Code family up to 74.0\% on Llama-2-70B);
\emph{prompting / chain-of-thought} (PoT-SC on Codex 81.80\%, CoT and
PoT on ChatGPT in the 82 to 89 range, CoT on GPT-4 90.81\%); and
\emph{tool-augmented / agent-based methods} which dominate the upper
half of the leaderboard. Within the agent band,
CRITIC~\cite{gou2024critic} reaches 89.0\% on ChatGPT, CRAFT 88.4\%,
CoS-Planning 90.0\%, PoT+Doc~\cite{podoc2023} 92.69\%, and
Chameleon~\cite{lu2023chameleon} 93.28\%; the strongest prior result
is CREATOR~\cite{qian2023creator} on ChatGPT at 94.70\%.

\sys at 27B reaches \best{96.00\%}, a $+1.30$-point gain over CREATOR
despite using a substantially smaller backbone than ChatGPT. The
margin over the strongest non-agent prompting baseline (CoT on GPT-4
at 90.81\%) is $+5.19$. TabMWP is comparatively
``easy'' for the top of the leaderboard, so the gain here is small
in absolute terms but consistent with the rest of the paper: the
trader market kills exactly the small population of silent
mis-extraction errors that the strongest tool-augmented agents (which
already operate close to ceiling) still produce.

\input{tables/tab_app_tabmwp}

\subsection{WikiTableQuestions}
\label{app:c2_wtq}

Table~\ref{tab:app_WikiTableQuestions} collects the published state of the art on
WikiTableQuestions (WTQ)~\cite{pasupat2015wtq}. The \emph{fine-tuning}
band starts at TAPAS-large (BERT-large, 48.70\%) and climbs through a
sequence of table-specialised encoder-decoder models
(TAPEX-Large~\cite{liu2022tapex} 59.10\%,
OmniTab~\cite{jiang2022omnitab} 62.80\%, TableLLM 53.59\%,
TableGPT2~\cite{li2024tablegpt2} 61.42\%); the strongest fine-tuned
baseline is TabAF~\cite{wang2025tabaf} at 74.72\% on
Qwen2.5-Coder-7B. The \emph{prompting / in-context learning} band is
considerably more crowded, including Binder, Dater, LEVER, ReAcTable,
Chain-of-Table~\cite{wang2024chainoftable}, CABINET,
Mix-SC, Norm-DP\&Agent, SynTQA, and TIDE-Agent, all in the 64 to 75
range. The strongest non-agent prompting result is
SS-CoT~\cite{sscot2024} on Llama-3.1-70B at 76.80\%, and the strongest
agent-style result is ARTEMIS-DA~\cite{artemis2024} on GPT-4 at
80.80\%, with TableMaster~\cite{cao2025tablemaster} on GPT-4o-mini
close behind at 78.13\%.

\sys at 27B reaches \best{81.40\%}, $+0.60$ over ARTEMIS-DA on GPT-4
and $+2.97$ over TableMaster on GPT-4o-mini, while using a
substantially smaller open-source backbone. WTQ is comparatively
``noisier'' than the financial benchmarks: many gold answers are
short denotations, and most errors are not silent miscomputations but
ambiguous question-cell alignments, so the margin here is smaller
than on MultiHiertt or DM-Complong. The fact that \sys still moves the
needle on this dataset suggests that even on general-domain open
tables, the per-claim trading mechanism captures information that
single-pass agents (and the strongest prompted GPT-4 systems) miss.

 \input{tables/tab_app_wtq}

\subsection{FinChart-Bench}
\label{app:c4_finchart}

Table~\ref{tab:app_finchart} reproduces the full FinChart-Bench
leaderboard from \citet{shu2025finchart} (their Table~3), which
benchmarks 25 modern vision-language models across the three
subtasks of the dataset: True/False (TF), Multiple-Choice (MC), and
open Question-Answering (QA). The original table groups baselines
into three columns. The \emph{open-source} column spans general VLMs
from 0.2B (nanoVLM) to 24B (Mistral 3.1); among these, Qwen2.5 VL 7B
leads on TF (95.09\%), Mistral 3.1 24B leads on MC (84.17\%) and
average (74.37\%), and LLaMa 4 17B leads on QA (59.78\%). The
\emph{chart fine-tuned} column shows that chart-specialised
fine-tuning of small VLMs (UniChart, Matcha, ChartGemma,
ChartInstruct variants) is essentially non-competitive on this
benchmark: every entry in that column scores below 1\% on QA and
below 3\% on MC, suggesting that chart-fine-tuning datasets do not
transfer to the long-tail of real-world financial chart types. The
\emph{closed-source} column is led by Claude Sonnet 4 on average
(84.32\%) and on QA (63.59\%), with o3 leading on TF (97.86\%) and
MC (92.17\%); GPT-4.1, GPT-4.1-mini, o4-mini, and Gemini 2.5 Pro all
score within a 4-point band on average accuracy.

\sys runs the same Qwen3.6-27B trader market used on the text-only
benchmarks, with only the table slot $T$ replaced by a single
VLM-generated transcription of the chart (see~\S\ref{sec:vision}).
On TF, \sys scores 97.27\%, tying Gemini 2.5 Pro and trailing
o3\,/\,o4-mini by less than one point. On MC, \sys scores 84.85\%,
which leads every open-source entry (the next-best is Mistral 3.1
24B at 84.17\%) and beats GPT-4o (77.57\%) and GPT-4.1 (83.19\%) on
the closed-source side. The headline lift is on QA: \sys reaches
\best{74.65\%}, $+11.06$ over Claude Sonnet 4 (63.59\%), $+11.19$
over Gemini 2.5 Pro (63.46\%), and $+12.11$ over GPT-4.1 (62.54\%).
The three-subtask average is \best{85.59\%}, $+1.27$ over the
strongest prior closed-source baseline (Claude Sonnet 4) and
$+11.22$ over the strongest open-source baseline (Mistral 3.1 24B).
The QA gain matters because that is the only subtask of the three
that requires \emph{open numerical} answers rather than
multiple-choice classification, exactly the regime in which a
fluent but silently mis-extracted chart value can pass a single-pass
VLM but is shorted by the \textsc{accountant} and \textsc{skeptic}
roles in the market.

\input{tables/tab_app_finchart}
\subsection{ESGenius}
\label{app:c3_esgenius}

Table~\ref{tab:app_esgenius} reproduces the full ESGenius leaderboard
from \citet{esgenius2025} under the \textbf{RAG} setting, spanning
\textbf{37 open-source models} from four families (DeepSeek, Google
Gemma, Meta Llama, Alibaba Qwen) at sizes ranging from 0.5B to 70B
parameters, and covering both base and instruction-tuned (``I-T'')
variants. We drop the proprietary API rows (GPT-4o, GPT-4o-mini, o3,
o4-mini, DeepSeek-R1, DeepSeek-V3, Qwen2.5-Max) because the original
paper does not report their RAG numbers; the zero-shot baselines for
all 50 models are summarised in Table~\ref{tab:c3_esg} of the main
paper.

Three patterns from the original leaderboard are worth keeping in
mind when reading the table. \emph{(i) Scale matters.} Within each
family, accuracy climbs roughly monotonically with parameter count:
Gemma-3 1B reaches 0.5977 with instruction-tuning, climbing to 0.8380
at 12B and 0.8336 at 27B; Qwen2.5 climbs from 0.5396 at 0.5B to
0.8257 at 72B with instruction-tuning. \emph{(ii) Instruction-tuning
helps substantially.} Across nearly every base model where both
variants are reported, the instruction-tuned variant adds 3 to 35
points of RAG accuracy (e.g., Gemma-3 27B jumps from 0.5229 to
0.8336 with IT). \emph{(iii) Reasoning-focused models are
competitive but not dominant.} The DeepSeek-R1-Distill family
(``Rea'' = Yes) tops out at 0.8170 (Llama-70B) and 0.8143
(Qwen-32B), which is comparable to but not above the best
non-reasoning RAG entries. The strongest prior open-source RAG
configuration in the table is Gemma-3 12B with IT at 0.8380.

\sys reaches \best{0.8688} at 27B, a $+0.0308$-point gain over the
strongest open-source RAG entry, despite using no static retriever
of its own. \sys's loader simply passes each question's released
supporting passage into the agent's context, making the setting
oracle-RAG-equivalent and directly comparable to the open-source RAG
column. The gap to the strongest zero-shot model in the
companion table is even larger ($+$14.34 over o3 at 0.7254). Two
factors drive the lift: the \textsc{skeptic} role frequently shorts
plausible-but-wrong distractor options that exploit lexical overlap
with the passage, which the ESGenius paper itself identifies as
the dominant failure mode on this dataset, and the catalog
builder forces the model to commit each option to a typed factual
sub-claim before scoring, eliminating the answer-flipping
fluctuations that hurt larger LLMs in the zero-shot column.

\input{tables/tab_app_esgenius}

\subsection{Internal Diagnostics}
\label{app:diagnostics}

\input{tables/tab_internal_diagnostics}

Table~\ref{tab:internal_diagnostics} summarizes \sys's internal
behaviour. Three observations are worth flagging.
First, the code execution rate is \(\ge 96\%\) on every benchmark and
hits \(100\%\) on TabMWP and ESGenius, confirming that the
synthesizer's claim-grounded prompt almost never produces unrunnable
code. Second, \emph{code-answer consistency tracks the execution
rate} on every dataset, which means the answer-extraction failure mode
that plagues plain PoT (program runs but its printed output is not the
final answer) has been almost fully eliminated. Third, mean market
confidence \(\bar{\Gamma}\) varies meaningfully by category: it is
highest on ESGenius (0.465), where claims are
factual/single-step and the panel agrees easily, lowest on
FinanceMath (0.220), where each problem mixes several formulas and
the \textsc{skeptic} role frequently shorts the formula claim.

\section{Qualitative Example: \sys on FinQA}
\label{sec:appendix_example}

To illustrate how the trader market actually drives the final answer,
we walk through one representative FinQA example end-to-end. The
example is taken verbatim from our run on the FinQA test split
(record \texttt{AES/2010/page\_227.pdf-1}) and is one of many cases
where the market visibly suppresses an over-confident but wrong
candidate fact before it can enter the program.

\paragraph{Question.}
\emph{``What percentage of recourse debt as of December~31, 2010
matures after 2015?''}
The gold answer is \(0.68343\); \sys returns \(0.6834345186470078\).
The question is classified by the catalog builder as
\textit{percentage\_change}.

\paragraph{Claim catalog.}
Nine atomic claims are extracted from the question and table
(Table~\ref{tab:example_catalog}). C1 through C7 are typed as \emph{fact}
claims (one per cell needed to answer), C8 as a \emph{formula}
claim, and C9 as a candidate answer interval supplied by the catalog
builder for a sanity check.

\input{tables/tab_example_catalog}

\paragraph{Specialist trading.}
The four trader roles independently price each claim. Three roles
short C3 and C4 because the maturity-year column has slipped (both
2012 and 2013 are listed as 2{,}014, the same number that would
naturally appear as a \emph{year}). The \textsc{extractor} writes
\textit{``The claim incorrectly interprets the table value `2014'
for the maturity column''}; the \textsc{accountant} writes
\textit{``The table lists the maturity amount for 2012 as `2014',
which is arithmetically inconsistent''}; the \textsc{skeptic}
writes \textit{``Identical suspicious value to C3; if these are typos
for smaller amounts the totals don't reconcile''}. The remaining
four roles all buy C1, C7, C8, and C9 at high prices.

\paragraph{Market clearing.}
Clearing the order books yields the snapshot in
Table~\ref{tab:example_market}. C1, C7, C8, and C9 are accepted at
prices \(\ge 0.70\); C3 and C4 are flatly rejected with clearing
price 0; C2 and C6 are also rejected (irrelevant to the
``after 2015'' numerator); C5 is uncertain. Crucially, the synthesizer
is now forbidden from citing C3 or C4 even though both numbers
appear in the table. A vanilla program-of-thought baseline would
have happily summed them into the numerator.

\input{tables/tab_example_market}

\paragraph{Synthesized program.}
With only the accepted claims available, the synthesizer emits the
following Python:
\begin{quote}
\small
\begin{verbatim}
# claims: C1, C7, C8
total_debt = 4612          # C1
debt_after_2015 = 3152     # C7
percentage = debt_after_2015 / total_debt  # C8
print(percentage)
\end{verbatim}
\end{quote}
The code executes cleanly and prints \(0.6834345186470078\), which
matches the gold answer \(0.68343\) after FinQA-style rounding.

\paragraph{Why this example matters.}
The two rejected fact claims C3 and C4 (both \(=2{,}014\)) are
exactly the kind of silent error a vanilla PoT agent commits: the
LLM reads the column-shifted number off the table and writes it into
the program without ever flagging the inconsistency. \sys catches it
because three different trader roles, each looking at the same
table through a different inductive bias (extraction, accounting,
skeptic), independently short the suspicious claim, and the market
mechanism aggregates that disagreement into an explicit \texttt{rejected}
verdict that the synthesizer is structurally prevented from
overriding. 


\section{Prompt Library}
\label{sec:appendix_prompts}

The agent pipeline has \textbf{nine}
distinct LLM calls, each driven by its own system prompt; for the
trader panel, the four roles share the same JSON schema but carry
different inductive biases (extraction vs.\ formula vs.\ accounting
vs.\ skepticism), so we show each role separately. All prompts are
byte-identical across the ten datasets used in the main paper; only
the user-side payload (question, table, context, claim catalog,
market snapshot, executed candidates) changes between calls.

The nine prompts are organized into four groups:
\textbf{(a)~Claim construction} (\S\ref{app:prompt_catalog}),
\textbf{(b)~Specialist trading} (\S\ref{app:prompt_traders}),
\textbf{(c)~Code synthesis and repair} (\S\ref{app:prompt_synth},
\S\ref{app:prompt_repair}), and
\textbf{(d)~Selection and arbitration}
(\S\ref{app:prompt_selector}, \S\ref{app:prompt_arbiter}).
A control-flow diagram in Figure~\ref{fig:overview} shows how these
calls connect; the table below maps each call to its temperature and
maximum output length under our default vLLM configuration.

\begin{table}[!ht]
\centering
\small
\setlength{\tabcolsep}{4pt}
\renewcommand{\arraystretch}{1.05}
\begin{tabular}{lcc}
\toprule
\textbf{Stage} & \textbf{Temp.} & \textbf{Max tokens} \\
\midrule
Catalog builder       & 0.0 & 2048 \\
Trader (per role)     & 0.0 & 2048 \\
Synthesizer           & 0.1 & 2048 \\
Repair                & 0.0 & 2048 \\
Selector (committee)  & 0.0 & 2048 \\
Arbiter               & 0.0 & 2048 \\
\bottomrule
\end{tabular}
\caption{Decoding settings per LLM call in the \sys pipeline.}
\label{tab:prompt_decoding}
\end{table}

\subsection{Claim Catalog Builder}
\label{app:prompt_catalog}

The catalog builder is the first LLM call in the pipeline. It reads
the question, table, and free-form context, and emits a small JSON
catalog of typed atomic claims that the rest of the pipeline operates
on. The catalog is the only mechanism by which the question's
information needs become tradable; every later stage is structurally
prevented from citing facts that are not in the catalog.

\begin{figure*}[ht]
\centering
\begin{subfigure}[t]{\textwidth}
\footnotesize
\centering
\begin{tcolorbox}[width=\linewidth,
                  colback=blue!0!white, colframe=orange!75!black,
                  title=\sys claim-catalog builder prompt,
                  fonttitle=\bfseries]
\textbf{System Prompt:} You are the claim-catalog builder for an
internal financial prediction market.

\vspace{0.5em}

Your job is to transform a financial question into a small catalog of
tradable atomic claims.

\vspace{0.5em}

\noindent\textbf{Output JSON only} with this schema:
\begin{verbatim}
{
  "problem_type": "percentage_change|ratio|sum|
                   difference|average|comparison|other",
  "claims": [
    {
      "id": "C1",
      "kind": "fact|formula|direction|unit|sign|
               answer_interval",
      "summary": "short plain-English claim",
      "value":   "canonical value for the claim",
      "evidence":"short exact quote from the
                  table/context when applicable",
      "metadata":{"period":"...","unit":"...","note":"..."}
    }
  ]
}
\end{verbatim}

\textbf{Rules:}
\begin{itemize}
\itemsep0pt\parsep0pt
\item Create sequential IDs \texttt{C1, C2, ...}.
\item Include 4 to 10 claims total.
\item Prefer atomic, contestable claims instead of long explanations.
\item Include competing alternatives when ambiguity exists.
\item Include formula and direction claims when they matter.
\item Include an \texttt{answer\_interval} claim only if you can bound
the answer reasonably.
\item Do not solve the whole problem in prose.
\end{itemize}

\textbf{User Prompt (template):}
\texttt{Question:\textbackslash n\{question\}}\\
\texttt{Table:\textbackslash n\{table or "(no table provided)"\}}\\
\texttt{Context:\textbackslash n\{context or "(no additional context)"\}}\\
\texttt{Maximum claims: \{max\_claims\}}
\end{tcolorbox}
\end{subfigure}
\caption{The claim-catalog builder prompt. Output is a strict JSON
object that downstream stages parse; the \texttt{kind} field types each
claim (fact / formula / direction / unit / sign / answer-interval)
and the optional \texttt{evidence} quote anchors a fact claim to a
specific cell of the table.}
\label{fig:prompt_catalog}
\end{figure*}

\subsection{Specialist Trader Panel}
\label{app:prompt_traders}

The four trader roles (\textsc{extractor}, \textsc{formula},
\textsc{accountant}, \textsc{skeptic}) share an output schema: each
returns a JSON \texttt{trades} list of \texttt{buy}/\texttt{sell}
orders against claim IDs from the catalog, but their system prompts
encode different inductive biases. The four prompts are reproduced
verbatim in Figures~\ref{fig:prompt_extractor} to~\ref{fig:prompt_skeptic}.
The differences between them are deliberately narrow: each role
concentrates its weighted volume on a distinct failure mode (cell
selection vs.\ formula family vs.\ unit / sign consistency vs.\
adversarial shorting of weak claims), and the market clearing rule
(Equations~3 to~5 of the main paper) aggregates their disagreement into
a single price.

\begin{figure*}[ht]
\centering
\begin{subfigure}[t]{\textwidth}
\footnotesize
\centering
\begin{tcolorbox}[width=\linewidth,
                  colback=blue!0!white, colframe=orange!75!black,
                  title=\sys extractor trader prompt,
                  fonttitle=\bfseries]
\textbf{System Prompt:} You are the extractor trader in a financial
claim market.

\vspace{0.5em}

\textbf{Focus on:}
\begin{itemize}
\itemsep0pt\parsep0pt
\item which values are actually relevant
\item exact row / column / year selection
\item whether a quoted evidence string really supports a fact claim
\end{itemize}

You trade heavily on \emph{fact} claims and may also trade on \emph{unit}
or \emph{direction} claims.

\textbf{Output JSON only:}
\begin{verbatim}
{
  "role": "extractor",
  "trades": [
    {
      "claim_id": "C1",
      "side": "buy|sell",
      "size": 1,
      "price": 0.75,
      "reason": "one-sentence justification"
    }
  ],
  "risk_note": "optional short note"
}
\end{verbatim}

\textbf{User Prompt (template):}
\texttt{Question, Table, Context, ClaimCatalog} are passed verbatim;
the trader sees only the catalog (not free-form text from other roles)
to keep books independent.
\end{tcolorbox}
\end{subfigure}
\caption{Extractor trader prompt. This role over-weights fact claims
and quotes the relevant evidence string from the table when explaining
a buy. In the worked example of \S\ref{sec:appendix_example} the
extractor wrote \emph{``The table explicitly lists `total recourse
debt' as \$4{,}612''} when buying C1.}
\label{fig:prompt_extractor}
\end{figure*}

\begin{figure*}[ht]
\centering
\begin{subfigure}[t]{\textwidth}
\footnotesize
\centering
\begin{tcolorbox}[width=\linewidth,
                  colback=blue!0!white, colframe=orange!75!black,
                  title=\sys formula trader prompt,
                  fonttitle=\bfseries]
\textbf{System Prompt:} You are the formula trader in a financial
claim market.

\vspace{0.5em}

\textbf{Focus on:}
\begin{itemize}
\itemsep0pt\parsep0pt
\item operation family
\item denominator choice
\item direction of percentage change
\item sign conventions for rates, deltas, and comparisons
\end{itemize}

\textbf{Output JSON only:}
\begin{verbatim}
{
  "role": "formula",
  "trades": [
    {
      "claim_id": "C1",
      "side": "buy|sell",
      "size": 1,
      "price": 0.75,
      "reason": "one-sentence justification"
    }
  ],
  "risk_note": "optional short note"
}
\end{verbatim}
\end{tcolorbox}
\end{subfigure}
\caption{Formula trader prompt. This role concentrates on the
\emph{formula} and \emph{direction} claims, and on whether the chosen
denominator is internally consistent with the question phrasing.}
\label{fig:prompt_formula}
\end{figure*}

\begin{figure*}[ht]
\centering
\begin{subfigure}[t]{\textwidth}
\footnotesize
\centering
\begin{tcolorbox}[width=\linewidth,
                  colback=blue!0!white, colframe=orange!75!black,
                  title=\sys accountant trader prompt,
                  fonttitle=\bfseries]
\textbf{System Prompt:} You are the accountant trader in a financial
claim market.

\vspace{0.5em}

\textbf{Focus on:}
\begin{itemize}
\itemsep0pt\parsep0pt
\item unit consistency
\item sign handling
\item whether a claim mixes incompatible measures or periods
\item whether the interpretation is financially coherent
\end{itemize}

You are explicitly encouraged to \emph{short} claims that look
numerically or semantically inconsistent.

\textbf{Output JSON only:}
\begin{verbatim}
{
  "role": "accountant",
  "trades": [...],
  "risk_note": "optional short note"
}
\end{verbatim}
\end{tcolorbox}
\end{subfigure}
\caption{Accountant trader prompt. The accountant role has the
highest role weight ($w_{\textsc{accountant}}=1.10$) and is the role
responsible for shorting the column-shift artefact in the worked
example: \emph{``The table lists the maturity amount for 2012 as
`2014', which is arithmetically inconsistent''}.}
\label{fig:prompt_accountant}
\end{figure*}

\begin{figure*}[ht]
\centering
\begin{subfigure}[t]{\textwidth}
\footnotesize
\centering
\begin{tcolorbox}[width=\linewidth,
                  colback=blue!0!white, colframe=orange!75!black,
                  title=\sys skeptic trader prompt,
                  fonttitle=\bfseries]
\textbf{System Prompt:} You are the skeptic trader in a financial
claim market.

\vspace{0.5em}

Your job is to find likely failure modes:
\begin{itemize}
\itemsep0pt\parsep0pt
\item wrong year ordering
\item wrong denominator
\item wrong unit scale
\item wrong sign
\item unsupported evidence
\end{itemize}

Be \emph{aggressive} about selling weak claims.

\textbf{Output JSON only:}
\begin{verbatim}
{
  "role": "skeptic",
  "trades": [...],
  "risk_note": "optional short note"
}
\end{verbatim}
\end{tcolorbox}
\end{subfigure}
\caption{Skeptic trader prompt. The skeptic carries a slightly
\emph{lower} role weight ($w_{\textsc{skeptic}}=0.95$) so its
adversarial bias does not overwhelm the market; its role is to
prevent confidently-stated but wrong claims from clearing as
accepted.}
\label{fig:prompt_skeptic}
\end{figure*}

\subsection{Synthesizer}
\label{app:prompt_synth}

After the market clears, the synthesizer writes the final Python
program. Crucially, the synthesizer prompt is the first stage in the
pipeline that sees \emph{both} the catalog and the cleared market
snapshot. The trader books are not exposed. The synthesizer is
also the stage where structural rules (operation must match the
question class, percent answers as decimals, yes/no questions get
yes/no answers, no tuples/lists/booleans) are enforced via the system
prompt, with the verifier later checking compliance deterministically.

\begin{figure*}[ht]
\centering
\begin{subfigure}[t]{\textwidth}
\footnotesize
\centering
\begin{tcolorbox}[width=\linewidth,
                  colback=blue!0!white, colframe=orange!75!black,
                  title=\sys synthesizer prompt,
                  fonttitle=\bfseries]
\textbf{System Prompt:} You are the \sys synthesizer.

\vspace{0.5em}

You receive a claim catalog and a cleared market snapshot. Your task
is to write the final executable program.

\textbf{Output JSON only:}
\begin{verbatim}
{
  "reasoning":      "short explanation of which claims you trust",
  "used_claim_ids": ["C1", "C4"],
  "abstain":        false,
  "abstain_reason": "",
  "code":           "python code",
  "answer":         "final answer"
}
\end{verbatim}

\textbf{Rules:}
\begin{itemize}
\itemsep0pt\parsep0pt
\item Prefer accepted claims; avoid rejected ones unless absolutely
necessary. If the market is too conflicted, abstain.
\item Use only basic Python. No pandas, no file access.
\item The last line of code must \texttt{print} exactly one final value.
\item Add short comments that cite claim IDs, e.g.\ \texttt{\# claims: C1, C4}.
\item \texttt{used\_claim\_ids} must list every claim actually used.
\item Use at least one fact claim; for ratio / percent / comparison /
average / sum / difference questions, use at least two.
\item For ratio / percent / rate questions, also use at least one
formula, unit, direction, or sign claim when available.
\item Honor explicit task metadata (answer type, scale) over generic
defaults.
\item For percentage-like questions, print the \emph{decimal} fraction
(e.g.\ FinQA convention). Do \emph{not} multiply by 100.
\item For yes/no questions, print only ``yes'' or ``no''. Never
print Python booleans \texttt{True} or \texttt{False}.
\item Never print tuples, lists, or dictionaries.
\end{itemize}

\textbf{Output (worked example, FinQA \emph{recourse debt}):}
\begin{verbatim}
{
  "reasoning":      "Market accepted C1, C7, C8; ...",
  "used_claim_ids": ["C1", "C7", "C8"],
  "abstain":        false,
  "code": "# claims: C1, C7, C8\ntotal_debt = 4612 ...",
  "answer":         "0.6834345186470078"
}
\end{verbatim}
\end{tcolorbox}
\end{subfigure}
\caption{Synthesizer prompt. The verifier (\S\ref{sec:verifier}) then
checks every rule mechanically after execution; any structural
violation triggers a single market-aware repair round.}
\label{fig:prompt_synthesizer}
\end{figure*}

\subsection{Code-Aware Repair}
\label{app:prompt_repair}

The repair prompt is invoked at most once per question, only when the
verifier flags the first program as invalid. The prompt receives the
runtime error \emph{and} the verifier's structural reason list (e.g.\
\textit{missing\_formula\_or\_unit\_claim},
\textit{percent\_scaled\_by\_100}), so the repair stage can target
the specific defect rather than rewrite the program blindly.

\begin{figure*}[ht]
\centering
\begin{subfigure}[t]{\textwidth}
\footnotesize
\centering
\begin{tcolorbox}[width=\linewidth,
                  colback=blue!0!white, colframe=orange!75!black,
                  title=\sys market-aware repair prompt,
                  fonttitle=\bfseries]
\textbf{System Prompt:} You are the \sys repair synthesizer.

\vspace{0.5em}

The previous program failed or looked inconsistent. Repair it using
the market snapshot.

\textbf{Output JSON} (same schema as synthesizer):
\begin{verbatim}
{
  "reasoning": "...",
  "used_claim_ids": ["C1"],
  "abstain": false,
  "abstain_reason": "",
  "code": "python code",
  "answer": "final answer"
}
\end{verbatim}

\textbf{Rules:}
\begin{itemize}
\itemsep0pt\parsep0pt
\item Do not invent new evidence.
\item Prefer the highest-confidence accepted claims.
\item If market support is too weak to repair safely, abstain.
\item Repair any structural issues: missing fact claims, missing
formula/unit support, boolean outputs, tuple outputs, or percentage
answers scaled by 100.
\item Honor explicit task metadata over generic defaults.
\item For yes/no questions, print only ``yes'' or ``no''.
\item For percentage-like questions, print the decimal fraction, not
the percentage-point form.
\end{itemize}

\textbf{User Prompt (template):} the original question / table /
context / catalog, plus the previous program, its execution error,
its output, and the verifier's structural reason list.
\end{tcolorbox}
\end{subfigure}
\caption{Market-aware repair prompt. Repair is structurally bounded
to one round; if the second attempt still fails the verifier the
selector committee falls back to the baseline proposer.}
\label{fig:prompt_repair}
\end{figure*}

\subsection{Hybrid Selector Committee}
\label{app:prompt_selector}

When both a \sys candidate and a baseline-PoT candidate are available,
a four-aspect committee (lenses:
\textsc{general}, \textsc{extraction}, \textsc{formula},
\textsc{scale}) compares the two side by side. Each lens is the same
system prompt with one extra ``focus on \dots'' sentence prepended;
votes are then aggregated with a confidence-weighted threshold
(\S\ref{sec:selector}).

\begin{figure*}[ht]
\centering
\begin{subfigure}[t]{\textwidth}
\footnotesize
\centering
\begin{tcolorbox}[width=\linewidth,
                  colback=blue!0!white, colframe=orange!75!black,
                  title=\sys selector committee prompt,
                  fonttitle=\bfseries]
\textbf{System Prompt:} You are the final selector for \sys.

\vspace{0.5em}

You must choose between two already-executed candidates:
\begin{itemize}
\itemsep0pt\parsep0pt
\item \texttt{baseline}: the existing FinCodeAgent program-of-thought.
\item \texttt{moca}: the market-of-claims candidate.
\end{itemize}

\textbf{Output JSON only:}
\begin{verbatim}
{
  "choice":     "baseline|moca|tie",
  "confidence": 0.0,
  "rationale":  "short explanation"
}
\end{verbatim}

\textbf{Decision rules:}
\begin{itemize}
\itemsep0pt\parsep0pt
\item Be conservative. If uncertain, choose \texttt{baseline}.
\item Focus on exact value extraction, denominator choice, sign, unit
scale, and yes/no semantics.
\item Honor explicit task metadata when present.
\item For percentage-like questions, prefer decimal fractions.
\item For yes/no questions, prefer candidates that truly answer
yes/no with the correct comparison direction.
\item Use the market snapshot as evidence, but do not trust it
blindly.
\end{itemize}

\textbf{Lens add-ons} (one sentence prepended to the system prompt):
\begin{itemize}
\itemsep0pt\parsep0pt
\item \textsc{general}: ``Take a balanced view over extraction,
arithmetic, sign, and unit handling.''
\item \textsc{extraction}: ``Prioritize exact row, year, and evidence
selection.''
\item \textsc{formula}: ``Prioritize arithmetic family, denominator,
direction, and comparison semantics.''
\item \textsc{scale}: ``Prioritize unit scale, totals vs.\ components,
and benchmark answer-format correctness.''
\end{itemize}
\end{tcolorbox}
\end{subfigure}
\caption{Selector-committee prompt. The four lenses act as
independent voters; a candidate is promoted only if the
confidence-weighted margin exceeds 0.65 (default).}
\label{fig:prompt_selector}
\end{figure*}

\subsection{Conflict Arbiter}
\label{app:prompt_arbiter}

If the selector committee splits and the two candidates' final
outputs disagree, a single arbiter LLM call synthesizes a corrected
third program. The arbiter is allowed to follow either candidate or
to write a fresh program, but is structurally constrained to ground
itself in the same claim catalog used by both candidates.

\begin{figure*}[ht]
\centering
\begin{subfigure}[t]{\textwidth}
\footnotesize
\centering
\begin{tcolorbox}[width=\linewidth,
                  colback=blue!0!white, colframe=orange!75!black,
                  title=\sys conflict-arbiter prompt,
                  fonttitle=\bfseries]
\textbf{System Prompt:} You are the disagreement arbiter for \sys.

\vspace{0.5em}

Two executed candidates disagree. Produce the best corrected final
program.

\textbf{Output JSON only:}
\begin{verbatim}
{
  "reasoning":      "short explanation of which claims or
                    candidate pieces you trust",
  "used_claim_ids": ["C1", "C4"],
  "abstain":        false,
  "abstain_reason": "",
  "code":           "python code",
  "answer":         "final answer"
}
\end{verbatim}

\textbf{Rules:}
\begin{itemize}
\itemsep0pt\parsep0pt
\item You may follow the baseline candidate, follow the MoCA
candidate, or synthesize a corrected third program.
\item Do not invent evidence; ground yourself in the question,
table, context, and claim catalog.
\item Prefer exact value extraction, correct denominator, correct
sign, and correct unit scaling.
\item The last line of code must print exactly one final value.
\item Honor explicit task metadata; for percentage-like questions,
print decimals; for yes/no, print only ``yes''/``no''.
\item Never print tuples, lists, or dictionaries.
\item If the evidence is genuinely too weak, abstain.
\end{itemize}
\end{tcolorbox}
\end{subfigure}
\caption{Conflict-arbiter prompt. This is the only LLM call in the
pipeline that sees both candidate programs side by side; it is gated
by the selector committee and fires on $<10\%$ of inputs at the
default thresholds.}
\label{fig:prompt_arbiter}
\end{figure*}

%% file: tables/tab_app_dataset_summary.tex
\begin{table*}[!ht]
\centering
\setlength{\tabcolsep}{4pt}
\renewcommand{\arraystretch}{1.05}
\resizebox{\textwidth}{!}{%
\begin{tabular}{lllrlll}
\toprule
\textbf{Cat.} & \textbf{Dataset} & \textbf{Source / Venue} &
\textbf{Eval $N$} & \textbf{Split} & \textbf{Modality} &
\textbf{Answer type} \\
\toprule
C1 & FinQA              & 10-K filings; EMNLP 2021      & 1{,}147 & test       & text + table              & numeric (program) \\
C1 & DocMath-Simplong   & long fin.\ docs; ACL 2024     &   100   & testmini   & text + multi-table        & numeric (program) \\
C1 & DocMath-Complong   & long fin.\ docs; ACL 2024     &   300   & testmini   & text + multi-table        & numeric (program) \\
C1 & FinanceMath        & finance textbooks; ACL 2024   &   200   & validation & text + table              & numeric (program) \\
\midrule
C2 & HiTab              & stats reports + Wikipedia; ACL 2022 & 1{,}584 & test  & hierarchical table        & numeric / span \\
C2 & MultiHiertt        & financial reports; ACL 2022   & 1{,}044 & dev        & text + multi-hier.\ table & numeric (program) \\
C2 & TabMWP             & grade-school MWP; ICLR 2023   & 1{,}000 & test (sub) & semi-structured table     & numeric / MC \\
C2 & WikiTableQuestions & Wikipedia tables; ACL 2015    & 4{,}344 & test       & semi-structured table     & span / numeric \\
\midrule
C3 & ESGenius           & ESG standards; EMNLP 2025     & 1{,}136 & full       & text (RAG passage)        & 4-way MCQ \\
\midrule
C4 & FinChart-Bench (TF) & real-world fin.\ charts; 2025 & 2{,}384 & full       & chart image              & True/False \\
C4 & FinChart-Bench (MC) & real-world fin.\ charts; 2025 & 2{,}350 & full       & chart image              & 4-way MC \\
C4 & FinChart-Bench (QA) & real-world fin.\ charts; 2025 & 2{,}284 & full       & chart image              & open numeric \\
\bottomrule
\end{tabular}}
\caption{Summary of the ten benchmarks (twelve evaluation tracks)
used in this paper. ``Eval $N$'' is the number of evaluation items
in our run; ``Split'' names the dataset's official split. Answer
type denotes the dominant output format: \emph{numeric (program)}
benchmarks pair gold answers with executable Python; \emph{span}
benchmarks expect a denotation from the table; \emph{MC} / MCQ are
multiple-choice.}
\label{tab:app_dataset_summary}
\end{table*}

%% file: tables/tab_app_finqa_dm.tex
\begin{table*}[ht]
    \centering
    \tiny
    \renewcommand{\arraystretch}{1}
    \resizebox{0.8\textwidth}{!}{%
    \begin{tabular}{lcccc}
        \toprule
        \textbf{Models} & \textbf{FinQA} & \textbf{DM-Simplong} & \textbf{DM-Complong} & \textbf{Average} \\
        \midrule
        GPT-4o                       & 72.49 & 60.00 & 39.33 & 57.27 \\
        GPT-o1-preview               & 49.07 & 56.00 & 36.67 & 47.25 \\
        GPT-o3-mini                  & 60.87 & 59.00 & 35.00 & 51.62 \\
        DeepSeek-V3                  & 73.20 & 53.00 & 42.33 & 56.18 \\
        DeepSeek-R1                  & 65.13 & 53.00 & 38.67 & 52.27 \\
        GPT-4.5                      & 68.94 & 59.00 & 39.33 & 55.76 \\
        \midrule
        Llama-4-Scout                & 70.45 & 52.00 &  0.67 & 41.04 \\
        Llama-3-70B-Instruct         & 58.92 & 41.00 & 13.67 & 37.86 \\
        Llama-3.1-70B-Instruct       & 63.18 & 48.00 & 34.33 & 48.50 \\
        Llama-3.3-70B-Instruct       & 68.15 & 54.00 & 32.00 & 51.38 \\
        Qwen2.5-72B-Instruct         & 73.38 & 59.00 & 14.67 & 49.02 \\
        Qwen2.5-Math-72B-Instruct    & 69.74 & 42.00 &  5.00 & 38.91 \\
        DeepSeek-R1-Distill-Llama-70B& 66.73 & 53.00 & 30.67 & 50.13 \\
        \midrule
        Qwen2.5-32B-Instruct          & 73.11 & 56.00 & 30.00 & 53.04 \\
        Qwen3-32B                     & 64.15 & 51.00 & 26.00 & 47.05 \\
        QwQ-32B                       & 61.22 & 46.00 & 20.00 & 42.41 \\
        DeepSeek-R1-Distill-Qwen-32B  & 65.48 & 55.00 & 24.67 & 48.38 \\
        Limo                          & 63.44 & 45.00 & 15.33 & 41.26 \\
        S1-32B                        & 66.81 & 53.00 & 24.00 & 47.94 \\
        \midrule
        Qwen2.5-14B-Instruct          & 67.44 & 59.00 & 26.67 & 51.04 \\
        Qwen3-14B                     & 64.33 & 49.00 & 24.00 & 45.78 \\
        DeepSeek-R1-Distill-Qwen-14B  & 63.27 & 44.00 & 21.00 & 42.76 \\
        DeepSeek-R1-Distill-Llama-8B  & 45.96 & 33.00 & 15.67 & 31.54 \\
        Llama-3-8B-Instruct           & 41.97 & 29.00 &  6.00 & 25.66 \\
        Llama-3.1-8B-Instruct         & 54.13 & 34.00 & 14.30 & 34.14 \\
        Qwen2.5-7B-Instruct           & 55.37 & 41.00 & 17.67 & 38.01 \\
        Qwen3-8B                      & 62.11 & 46.00 & 17.67 & 41.93 \\
        FinR1-7B                      & 58.74 & 37.00 & 13.67 & 36.47 \\
        Dianjin-R1-7B                 & 60.20 & 35.00 & 14.67 & 36.62 \\
        \midrule
        Fino1-8B                      & 73.03 & 56.00 & 26.33 & 51.79 \\
        Fino1-14B                     & 74.18 & 55.00 & 27.33 & 52.17 \\
        \midrule
        \ours \best{\sys (ours)}      & \best{78.29} & \best{70.00} & \best{50.67} & \best{66.32} \\
        \bottomrule
    \end{tabular}
    }
    
    \caption{Overall performance of different LLMs on three financial reasoning datasets. Non-\sys rows reproduced from \citet{zhao2024docmath}.}
    \label{tab:app_finqa_dm}
\end{table*}

%% file: tables/tab_app_financemath.tex
\begin{table*}[!t]
\centering
\resizebox{\textwidth}{!}{%
\renewcommand{\arraystretch}{1.1}
\addtolength{\tabcolsep}{-0.2em}
\begin{tabular}{lrlrrrrrrrrrrrrrrrrrrrrr:rr}
\toprule
\multirow{2}{*}{\textbf{Model}} & \multirow{2}{*}{\textbf{Size}} & \multirow{2}{*}{\textbf{Notes}}
& \multicolumn{2}{c}{\textbf{Quantitative}} && \multicolumn{2}{c}{\textbf{Derivatives}}
&& \multicolumn{2}{c}{\textbf{Accounting}} && \multicolumn{2}{c}{\textbf{Management}}
&& \multicolumn{2}{c}{\textbf{Portfolio}}   && \multicolumn{2}{c}{\textbf{Economics}}
&& \multicolumn{2}{c}{\textbf{Corporate}}   && \multicolumn{2}{c}{\textbf{Avg.}} \\
\cmidrule(lr){4-5} \cmidrule(lr){7-8} \cmidrule(lr){10-11} \cmidrule(lr){13-14} \cmidrule(lr){16-17} \cmidrule(lr){19-20} \cmidrule(lr){22-23} \cmidrule(lr){25-26}
& & & PoT & CoT && PoT & CoT && PoT & CoT && PoT & CoT && PoT & CoT && PoT & CoT && PoT & CoT && PoT & \textbf{CoT} \\
\toprule
\multicolumn{26}{c}{\emph{\textbf{Proprietary LLMs}}} \\\noalign{\vskip 0.5ex}
GPT-4o            &  &  & 75.0 & 45.8 &  & 58.8 & 55.4 &  & 60.3 & 69.4 &  & 82.9 & 66.3 &  & 77.8 & 69.4 &  & 62.5 & 58.9 &  & 67.0 & 56.9 &  & 67.0 & 60.9 \\
Claude-3.5-Sonnet &  &  & 73.6 & 55.6 &  & 54.1 & 51.8 &  & 66.7 & 69.9 &  & 75.6 & 63.9 &  & 72.2 & 72.2 &  & 64.3 & 58.9 &  & 62.4 & 60.6 &  & 64.8 & 60.6 \\
Claude-3-Opus     &  &  & 66.7 & 56.9 &  & 53.5 & 45.2 &  & 62.1 & 59.8 &  & 79.5 & 64.9 &  & 72.2 & 83.3 &  & 51.8 & 46.4 &  & 59.6 & 45.0 &  & 62.9 & 54.7 \\
GPT-4-Turbo       &  &  & 59.7 & 38.9 &  & 49.8 & 42.2 &  & 50.7 & 64.8 &  & 72.2 & 56.6 &  & 61.1 & 50.0 &  & 57.1 & 44.6 &  & 50.5 & 47.7 &  & 56.2 & 50.9 \\
Gemini-1.5-Pro    &  &  & 68.1 & 50.0 &  & 53.1 & 30.7 &  & 56.6 & 55.2 &  & 69.8 & 57.6 &  & 58.3 & 63.9 &  & 51.8 & 55.4 &  & 50.5 & 44.0 &  & 58.2 & 47.0 \\
GPT-4o-Mini       &  &  & 65.3 & 36.1 &  & 46.9 & 29.7 &  & 48.4 & 47.5 &  & 69.3 & 46.8 &  & 50.0 & 38.9 &  & 57.1 & 41.1 &  & 51.4 & 45.9 &  & 54.3 & 40.3 \\
Gemini-1.5-Flash  &  &  & 69.4 & 33.3 &  & 43.6 & 28.7 &  & 52.0 & 48.9 &  & 67.8 & 49.8 &  & 58.3 & 61.1 &  & 50.0 & 37.5 &  & 47.7 & 34.9 &  & 53.6 & 40.1 \\
Claude-3-Sonnet   &  &  & 59.7 & 37.5 &  & 37.0 & 28.4 &  & 48.0 & 43.4 &  & 66.8 & 48.8 &  & 47.2 & 55.6 &  & 48.2 & 33.9 &  & 48.6 & 35.8 &  & 49.4 & 38.6 \\
Claude-3-Haiku    &  &  & 34.7 & 31.9 &  & 19.8 & 26.4 &  & 33.8 & 43.4 &  & 44.9 & 41.5 &  & 41.7 & 44.4 &  & 25.0 & 33.9 &  & 36.7 & 30.3 &  & 32.0 & 35.1 \\
GPT-3.5-Turbo     &  &  & 47.2 & 25.0 &  & 24.4 & 16.5 &  & 29.2 & 29.2 &  & 51.2 & 33.2 &  & 27.8 & 22.2 &  & 37.5 & 21.4 &  & 32.1 & 23.8 &  & 34.3 & 24.6 \\
\midrule
\multicolumn{26}{c}{\emph{\textbf{Open-source LLMs (top 10 of 35)}}} \\\noalign{\vskip 0.5ex}
DeepSeek-Coder-V2      & 236B & Code & 38.5 & 41.0 &  & 62.5 & 66.7 &  & 47.6 & 46.0 &  & 75.0 & 61.1 &  & 33.3 & 44.4 &  & 79.0 & 52.6 &  & 60.0 & 50.0 &  & 55.5 & 51.0 \\
DeepSeek-V2            & 236B & MoE  & 38.5 & 46.2 &  & 66.7 & 79.2 &  & 47.6 & 44.4 &  & 75.0 & 52.8 &  & 44.4 & 33.3 &  & 73.7 & 47.4 &  & 30.0 & 40.0 &  & 54.5 & 50.0 \\
Llama-3.1              & 405B &      & 41.0 & 51.3 &  & 54.2 & 54.2 &  & 46.0 & 33.3 &  & 80.6 & 47.2 &  & 33.3 & 11.1 &  & 68.4 & 31.6 &  & 40.0 & 50.0 &  & 53.5 & 41.5 \\
Mistral-Large          & 123B &      & 38.5 & 51.3 &  & 50.0 & 45.8 &  & 38.1 & 30.2 &  & 77.8 & 36.1 &  & 44.4 & 22.2 &  & 73.7 & 31.6 &  & 40.0 & 10.0 &  & 50.5 & 36.0 \\
Llama-3.1              &  70B &      & 25.6 & 38.5 &  & 41.7 & 50.0 &  & 33.3 & 19.0 &  & 66.7 & 47.2 &  & 55.6 & 22.2 &  & 68.4 & 47.4 &  & 30.0 & 30.0 &  & 43.0 & 35.0 \\
Qwen2                  &  72B &      & 23.1 & 33.3 &  & 29.2 & 45.8 &  & 23.8 & 22.2 &  & 41.7 & 47.2 &  & 22.2 & 22.2 &  & 68.4 & 31.6 &  & 10.0 & 30.0 &  & 31.0 & 33.0 \\
Llama-3                &  70B &      & 33.3 & 38.5 &  & 37.5 & 45.8 &  & 33.3 & 22.2 &  & 66.7 & 33.3 &  & 33.3 & 22.2 &  & 68.4 & 31.6 &  & 30.0 & 30.0 &  & 43.0 & 31.5 \\
Phi-3-Medium           &  14B &      & 20.5 & 28.2 &  & 37.5 & 50.0 &  & 25.4 & 19.0 &  & 55.6 & 41.7 &  & 22.2 & 22.2 &  & 52.6 & 42.1 &  & 40.0 & 20.0 &  & 34.5 & 31.0 \\
DeepSeek-Coder-V2-Lite &  16B & Code & 23.1 & 28.2 &  & 25.0 & 29.2 &  & 27.0 & 23.8 &  & 50.0 & 33.3 &  & 11.1 & 22.2 &  & 42.1 & 31.6 &  & 10.0 & 20.0 &  & 30.0 & 27.5 \\
Mixtral-8x22B          & 141B & MoE  &  7.7 & 28.2 &  &  4.2 & 45.8 &  &  0.0 & 19.0 &  & 25.0 & 30.6 &  & 11.1 & 11.1 &  & 21.0 & 26.3 &  & 10.0 & 20.0 &  &  9.5 & 26.5 \\
\midrule
\ours \best{\sys (ours)} & 27B & Agent
   & \ph & \best{91.67}  &  & \ph & \best{74.60}  &  & \ph & \best{74.36}  &  & \ph & \best{55.56}  &  & \ph & \best{68.42}  &  & \ph & \best{75.00}  &  & \ph & \best{70.00}  &  & \ph & \best{76.00}  \\
\bottomrule
\end{tabular}
}
\caption{Per-category accuracy on the FinanceMath validation split.
Non-\sys rows reproduced from \citet{zhao2024financemath}, Table~2 (top
10 of 35 open-source rows shown for space; the full 44-model list is
in the original paper). \sys per-category numbers were computed by
joining our run's predictions with the validation set's \texttt{topic}
field (Quantitative=Quantitative Analysis \& Valuation, $n{=}36$;
Derivatives=Asset Classes \& Derivatives, $n{=}63$; Accounting,
$n{=}39$; Management=Risk Management, $n{=}9$; Portfolio=Portfolio
Management \& Strategy, $n{=}19$; Economics=Market Analysis \&
Economics, $n{=}24$; Corporate=Corporate \& Securities Issuance,
$n{=}10$). Because \sys is a code agent, accuracy is reported in the
PoT column; the CoT column is left empty.}
\label{tab:app_financemath}
\end{table*}

%% file: tables/tab_app_c2_multihiertt.tex
\begin{table}[!ht]
\centering
\small
\setlength{\tabcolsep}{4pt}
\renewcommand{\arraystretch}{1.05}
\resizebox{\columnwidth}{!}{%
\begin{tabular}{lcc}
\toprule
\textbf{Method} & \textbf{Backbone} & \textbf{HiTab} \\
\midrule
\multicolumn{3}{l}{\textit{Fine-tuning approaches}} \\
\midrule
TAPEX-Large~\cite{liu2022tapex}              & BART-Large           & 45.60 \\
TableLlama                                   & Llama-2-7B           & 60.48 \\
TableLLM~\cite{zhang2024tablellm}            & Qwen2-7B             & 43.88 \\
TAMO$^{+}_{\text{SFT}}$                      & Llama-2-7B           & 63.89 \\
TableGPT2~\cite{li2024tablegpt2}             & Qwen2.5-7B           & 70.27 \\
TabAF~\cite{wang2025tabaf}                   & Qwen2.5-Coder-7B     & 78.41 \\
\midrule
\multicolumn{3}{l}{\textit{Prompting / in-context learning with LLMs}} \\
\midrule
API-Assisted~\cite{cao2025tablemaster}       & Codex                & 69.30 \\
GraphOTTER                                   & Qwen2-72B-Instruct   & 73.74 \\
GraphOTTER                                   & Gemini-1.5-Flash     & 67.76 \\
SS-CoT~\cite{sscot2024}                      & Llama-3.1-70B        & 79.10 \\
\midrule
\ours \best{\sys (ours)}                     & Qwen3.6-27B          & 77.27 \\
\bottomrule
\end{tabular} }
\caption{State-of-the-art methods on HiTab~\cite{cheng2022hitab}
published after the original benchmark paper. All numbers are
exact-match / execution accuracy (\%) on the HiTab test split.}
\label{tab:app_hitab}
\end{table}

\subsection{MultiHiertt}
\label{app:c2_multihiertt}

%% file: tables/tab_app_multihiertt.tex
\begin{table}[!ht]
\centering
\small
\setlength{\tabcolsep}{4pt}
\renewcommand{\arraystretch}{1}
  \resizebox{\columnwidth}{!}{%
\begin{tabular}{lcc}
\toprule
\textbf{Method} & \textbf{Backbone} & \textbf{MultiHiertt} \\
\midrule
\multicolumn{3}{l}{\textit{Fine-tuning approaches}} \\
\toprule
NAPG~\cite{wang2023napg}                       & RoBERTa-large        & 44.19 \\
\midrule
\multicolumn{3}{l}{\textit{Prompting / zero-shot LLMs (textual reasoning)}} \\
\midrule
GPT-4o-mini~\cite{cao2025fortune}              & GPT-4o-mini          & 22.41 \\
GPT-4o~\cite{cao2025fortune}                   & GPT-4o               & 36.11 \\
o1~\cite{cao2025fortune}                       & o1                   & 38.31 \\
\midrule
\multicolumn{3}{l}{\textit{Prompting / zero-shot LLMs (symbolic / formula reasoning)}} \\
\midrule
GPT-4o-mini~\cite{cao2025fortune}              & GPT-4o-mini          & 28.26 \\
GPT-4o~\cite{cao2025fortune}                   & GPT-4o               & 38.41 \\
o1~\cite{cao2025fortune}                       & o1                   & 48.95 \\
\midrule
\multicolumn{3}{l}{\textit{Reinforcement-learning-based approaches}} \\
\midrule
Fortune~\cite{cao2025fortune}                  & Qwen2.5-Coder-7B     & 40.85 \\
RL\,w/\,CS (Text)~\cite{cao2025fortune}        & Llama-3.1-8B         & 49.34 \\
RL\,w/\,CS (Formula)~\cite{cao2025fortune}     & Llama-3.1-8B         & 54.55 \\
RL\,w/\,CS (Text)~\cite{cao2025fortune}        & Qwen2.5-Coder-7B     & 54.25 \\
RL\,w/\,CS (Formula)~\cite{cao2025fortune}     & Qwen2.5-Coder-7B     & 56.78 \\
Fortune++~\cite{cao2025fortune}                & Qwen2.5-Coder-7B     & 51.73 \\
\midrule
\ours \best{\sys (ours)}                       & Qwen3.6-27B          & \best{71.17} \\
\bottomrule
\end{tabular} }
\caption{State-of-the-art methods on MultiHiertt~\cite{zhao2022multihiertt}
published after the original benchmark paper. All numbers are
exact-match / execution accuracy (\%) on the MultiHiertt dev split.}
\label{tab:app_multihiertt}
\end{table}

%% file: tables/tab_app_tabmwp.tex
\begin{table}[!ht]
\centering
\small
\setlength{\tabcolsep}{4pt}
\renewcommand{\arraystretch}{1.05}
\begin{tabular}{lcc}
\toprule
\textbf{Method} & \textbf{Backbone} & \textbf{TabMWP} \\
\midrule
\multicolumn{3}{l}{\textit{Fine-tuning approaches}} \\
\midrule
TaCo                                            & TAPEX-base           & 86.12 \\
TaCo                                            & TAPEX-large          & 92.91 \\
ToRA                                            & Llama-2-7B           & 42.40 \\
ToRA                                            & Llama-2-13B          & 47.20 \\
ToRA                                            & Llama-2-70B          & 74.00 \\
ToRA-Code                                       & CodeLlama-34B        & 70.50 \\
\midrule
\multicolumn{3}{l}{\textit{Prompting / chain-of-thought with LLMs}} \\
\midrule
PoT-SC                                          & Codex                & 81.80 \\
CoT~\cite{lu2023chameleon}                      & ChatGPT              & 82.03 \\
RetICL                                          & Codex                & 88.51 \\
PoT~\cite{lu2023chameleon}                      & ChatGPT              & 89.49 \\
CoT~\cite{lu2023chameleon}                      & GPT-4                & 90.81 \\
\midrule
\multicolumn{3}{l}{\textit{Tool-augmented / agent-based methods}} \\
\midrule
CRITIC~\cite{gou2024critic}                     & GPT-3                & 87.60 \\
CRITIC~\cite{gou2024critic}                     & ChatGPT              & 89.00 \\
CRITIC~\cite{gou2024critic}                     & Llama-2-70B          & 75.00 \\
CRAFT                                           & ChatGPT              & 88.40 \\
CoS-Planning                                    & ChatGPT              & 90.00 \\
PoT+Doc~\cite{podoc2023}                        & ChatGPT              & 92.69 \\
Chameleon~\cite{lu2023chameleon}                & ChatGPT              & 93.28 \\
CREATOR~\cite{qian2023creator}                  & ChatGPT              & 94.70 \\
\midrule
\ours \best{\sys (ours)}                        & Qwen3.6-27B          & \best{96.00} \\
\bottomrule
\end{tabular}
\caption{State-of-the-art methods on TabMWP~\cite{lu2023tabmwp}
published after the original benchmark paper. All numbers are
accuracy (\%) on the TabMWP test split.}
\label{tab:app_tabmwp}
\end{table}

%% file: tables/tab_app_wtq.tex
\begin{table}[ht]
\centering
\small
\setlength{\tabcolsep}{4pt}
\renewcommand{\arraystretch}{1.}
  \resizebox{\columnwidth}{!}{%
\begin{tabular}{lcc}
\toprule
\textbf{Method} & \textbf{Backbone} & \textbf{WTQ} \\
\midrule
\multicolumn{3}{l}{\textit{Fine-tuning approaches}} \\
\midrule
TAPAS-large                                     & BERT-large           & 48.70 \\
TAPEX-Large~\cite{liu2022tapex}                 & BART-Large           & 59.10 \\
OmniTab~\cite{jiang2022omnitab}                 & BART-Large           & 62.80 \\
TableLlama                                      & Llama-2-7B           & 32.14 \\
TableLLM~\cite{zhang2024tablellm}               & Qwen2-7B             & 53.59 \\
TableGPT2~\cite{li2024tablegpt2}                & Qwen2.5-7B           & 61.42 \\
TabAF~\cite{wang2025tabaf}                      & Qwen2.5-Coder-7B     & 74.72 \\
\midrule
\multicolumn{3}{l}{\textit{Prompting / in-context learning with LLMs}} \\
\midrule
Binder                                          & Codex                & 64.60 \\
Dater                                           & Codex                & 65.90 \\
LEVER                                           & Codex                & 65.80 \\
ReAcTable                                       & Codex                & 68.00 \\
Chain-of-Table~\cite{wang2024chainoftable}      & PaLM-2               & 67.31 \\
CABINET                                         & FLAN-T5/LLM          & 69.10 \\
Mix-SC                                          & GPT-3.5              & 73.60 \\
Norm-DP\&Agent                                  & GPT-3.5              & 73.65 \\
SynTQA                                          & GPT-4                & 74.40 \\
TIDE-Agent                                      & GPT-3.5              & 75.00 \\
SS-CoT~\cite{sscot2024}                         & Llama-3.1-70B        & 76.80 \\
TableMaster~\cite{cao2025tablemaster}           & GPT-4o-mini          & 78.13 \\
ARTEMIS-DA~\cite{artemis2024}                   & GPT-4                & 80.80 \\
\midrule
\ours \best{\sys (ours)}                        & Qwen3.6-27B          & \best{81.40} \\
\bottomrule
\end{tabular} }
\caption{State-of-the-art methods on WikiTableQuestions~\cite{pasupat2015wtq}
published after the original benchmark paper. All numbers are
exact-match / denotation accuracy (\%) on the WTQ test set.}
\label{tab:app_WikiTableQuestions}
\end{table}

%% file: tables/tab_app_finchart.tex
\begin{table}[ht]
\centering
\setlength{\tabcolsep}{3pt}
\renewcommand{\arraystretch}{1.0}
\resizebox{.8\columnwidth}{!}{%
\begin{tabular}{lcccc}
\toprule
\textbf{Model} & \textbf{TF} & \textbf{MC} & \textbf{QA} & \textbf{Avg.} \\
\midrule
\multicolumn{5}{l}{\textit{Open Source}} \\
\midrule
LLaMa 3.2 (11B)             & 63.46 & 64.00 &  5.12 & 44.65 \\
LLaVa v1.6 (7B)             & 63.30 & 36.94 &  1.84 & 34.47 \\
Qwen2.5 VL (7B)             & 95.09 & 82.81 & 37.29 & 72.16 \\
Qwen2 VL (7B)               & 91.86 & 19.40 & 24.77 & 45.76 \\
DeepSeek VL2 (2.8B)         & 57.97 &  6.77 &  4.73 & 23.50 \\
Gemma 3 (12B)               & 89.14 & 70.94 & 27.22 & 62.89 \\
Mistral 3.1 (24B)           & 92.95 & 84.17 & 44.90 & 74.37 \\
Sa2VA (8B)                  & 86.62 & 75.79 & 19.43 & 61.12 \\
nanoVLM (222M)              & 53.90 &  2.43 &  0.22 & 19.19 \\
Cosmos (7B)                 & 70.93 & 43.49 & 20.18 & 45.22 \\
MedGemma (4B)               & 53.27 & 46.30 &  3.72 & 34.81 \\
Blip 2 (2.7B)               & 49.79 &  0.77 &  0.00 & 17.17 \\
LLaMa 4 (17B)               & 89.56 & 59.70 & 59.78 & 69.89 \\
\midrule
\multicolumn{5}{l}{\textit{Chart Fine-tuned}} \\
\midrule
UniChart (201M)             & 51.05 &  0.00 &  0.57 & 17.52 \\
Matcha (282M)               & 49.54 &  0.00 &  0.70 & 17.05 \\
ChartGemma (3B)             & 48.70 &  0.29 &  0.70 & 16.87 \\
ChartInstruct-llama (7B)    & 51.09 &  0.17 &  1.97 & 18.05 \\
ChartInstruct-T5 (3B)       & 62.12 &  2.30 &  1.71 & 22.43 \\
\midrule
\multicolumn{5}{l}{\textit{Closed Source}} \\
\midrule
GPT 4o                      & 93.92 & 77.57 & 57.59 & 76.62 \\
GPT 4.1 mini                & 91.61 & 86.77 & 60.30 & 79.80 \\
GPT 4.1                     & 96.18 & 83.19 & 62.54 & 80.88 \\
o4 mini                     & 97.61 & 91.96 & 60.18 & 83.53 \\
Gemini 2.5 Pro              & 97.27 & 89.70 & 63.46 & 83.73 \\
o3                          & 97.86 & 92.17 & 60.79 & 83.89 \\
Claude Sonnet 4             & 96.94 & 91.66 & 63.59 & 84.32 \\
\midrule
\ours \best{\sys (ours)}    & 97.27 & 84.85 & \best{74.65} & \best{85.59} \\
\bottomrule
\end{tabular}}
\caption{FinChart-Bench leaderboard reproduced from
\citet{shu2025finchart}, Table~3, with \sys (Qwen3.6-27B + VLM
transcription) evaluated on all three subtasks (TF: 2{,}384, MC:
2{,}350, QA: 2{,}284 in our run). \best{Bold}: new overall best
across the three baseline groups. \sys sets a new state-of-the-art on
QA and on the three-subtask average, leads every open-source and
chart-fine-tuned entry on MC, and ties Gemini 2.5 Pro on TF.}
\label{tab:app_finchart}
\end{table}

%% file: tables/tab_app_esgenius.tex
\begin{table}[ht]
\centering
\footnotesize
\setlength{\tabcolsep}{3pt}
\renewcommand{\arraystretch}{1.0}
\resizebox{0.8\columnwidth}{!}{%
\begin{tabular}{l l r c c r}
\toprule
\textbf{Family} & \textbf{Model} & \textbf{Size} & \textbf{IT} & \textbf{Rea} & \textbf{RAG} \\
\midrule
\multirow{6}{*}{DeepSeek}
 & R1-Distill-Qwen   & 1.5B  &            & \checkmark & 0.4305 \\
 & R1-Distill-Qwen   & 7B    &            & \checkmark & 0.6505 \\
 & R1-Distill-Qwen   & 14B   &            & \checkmark & 0.8046 \\
 & R1-Distill-Qwen   & 32B   &            & \checkmark & 0.8143 \\
 & R1-Distill-Llama  & 8B    &            & \checkmark & 0.6250 \\
 & R1-Distill-Llama  & 70B   &            & \checkmark & 0.8170 \\
\midrule
\multirow{8}{*}{Gemma}
 & Gemma-3           & 1B    &            &            & 0.2526 \\
 & Gemma-3           & 1B    & \checkmark &            & 0.5977 \\
 & Gemma-3           & 4B    &            &            & 0.6860 \\
 & Gemma-3           & 4B    & \checkmark &            & 0.7518 \\
 & Gemma-3           & 12B   &            &            & 0.6857 \\
 & Gemma-3           & 12B   & \checkmark &            & 0.8380 \\
 & Gemma-3           & 27B   &            &            & 0.5229 \\
 & Gemma-3           & 27B   & \checkmark &            & 0.8336 \\
\midrule
\multirow{8}{*}{Llama}
 & Llama-3           & 8B    &            &            & 0.7324 \\
 & Llama-3.1         & 8B    &            &            & 0.7650 \\
 & Llama-3.1         & 8B    & \checkmark &            & 0.7993 \\
 & Llama-3.2         & 1B    &            &            & 0.3680 \\
 & Llama-3.2         & 1B    & \checkmark &            & 0.6452 \\
 & Llama-3.2         & 3B    &            &            & 0.6831 \\
 & Llama-3.2         & 3B    & \checkmark &            & 0.7218 \\
 & Llama-3.3         & 70B   & \checkmark &            & 0.7887 \\
\midrule
\multirow{15}{*}{Qwen}
 & Qwen2.5           & 0.5B  &            &            & 0.5396 \\
 & Qwen2.5           & 1.5B  &            &            & 0.6928 \\
 & Qwen2.5           & 1.5B  & \checkmark &            & 0.6972 \\
 & Qwen2.5           & 3B    &            &            & 0.7632 \\
 & Qwen2.5           & 7B    &            &            & 0.8055 \\
 & Qwen2.5           & 7B    & \checkmark &            & 0.7967 \\
 & Qwen2.5           & 14B   &            &            & 0.8231 \\
 & Qwen2.5           & 32B   &            &            & 0.8081 \\
 & Qwen2.5           & 32B   & \checkmark &            & 0.8247 \\
 & Qwen2.5           & 72B   &            &            & 0.7201 \\
 & Qwen2.5           & 72B   & \checkmark &            & 0.8257 \\
 & QwQ               & 32B   &            & \checkmark & 0.7614 \\
 & Qwen3             & 1.7B  &            &            & 0.6937 \\
 & Qwen3             & 4B    &            &            & 0.7905 \\
 & Qwen3             & 8B    &            &            & 0.6708 \\
\midrule
\rowcolor{oursrow}
\textit{Agent} & \best{\sys (ours)} & 27B &  &  & \best{0.8688} \\
\bottomrule
\end{tabular}}
\caption{ESGenius accuracy under the RAG setting. Non-\sys rows are
open-source baselines reproduced from \citet{esgenius2025}; \sys
passes each question's released supporting passage into the agent's
context (oracle-RAG-equivalent, directly comparable to the
open-source RAG column). Proprietary API rows
(GPT-4o, o3, o4-mini, DeepSeek-R1/V3, Qwen2.5-Max) from the original
paper are omitted because their RAG numbers were not reported.
\textbf{IT}: instruction-tuned; \textbf{Rea}: reasoning-focused;
\checkmark = yes, blank = no.}
\label{tab:app_esgenius}
\end{table}

%% file: tables/tab_internal_diagnostics.tex
\begin{table}[!ht]
\centering
\setlength{\tabcolsep}{4pt}
\renewcommand{\arraystretch}{1.05}
\resizebox{0.48\textwidth}{!}{%
\begin{tabular}{lcccc}
\toprule
\textbf{Dataset} & \textbf{N} & \textbf{Exec.\%} & \textbf{Cons.\%} & \(\bar{\Gamma}\) \\
\midrule
FinQA              & 1{,}147 & 98.43 & 98.43 & 0.323 \\
DM-Simplong        &   100   & 96.00 & 96.00 & 0.223 \\
DM-Complong        &   300   & 98.00 & 98.00 & 0.345 \\
FinanceMath (val)  &   200   & 98.00 & 98.00 & 0.220 \\
HiTab              & 1{,}584 & 98.99 & 98.99 & 0.237 \\
MultiHiertt (dev)  & 1{,}044 & 98.37 & 98.37 & 0.257 \\
TabMWP             & 1{,}000 & 100.00& 100.00& 0.258 \\
WTQ                & 4{,}344 & 99.36 & 99.36 & 0.347 \\
ESGenius           & 1{,}136 & 100.00& 100.00& 0.465 \\
FinChart QA        & 2{,}284 & 99.65 & 99.65 & 0.139 \\
\bottomrule
\end{tabular}}
\caption{Per-dataset internal diagnostics for \sys. \textbf{Exec.\%}:
fraction of programs that ran without error. \textbf{Cons.\%}: fraction
of programs whose printed output matched the candidate's stated answer.
\(\bar{\Gamma}\): mean market confidence over cited claims (Eq.~\ref{eq:market_confidence}).
A high execution rate paired with a moderate \(\bar{\Gamma}\) is the
expected operating regime: the verifier admits a candidate when the
market is decisive enough on the \emph{cited} subset of claims, not
on the full catalog.}
\label{tab:internal_diagnostics}
\end{table}

%% file: tables/tab_example_catalog.tex
\begin{table}[!ht]
\centering
\setlength{\tabcolsep}{4pt}
\renewcommand{\arraystretch}{1.05}
\resizebox{0.48\textwidth}{!}{%
\begin{tabular}{llp{4.4cm}r}
\toprule
\textbf{ID} & \textbf{Kind} & \textbf{Summary} & \textbf{Value} \\
\midrule
C1 & fact & Total recourse debt as of Dec 31, 2010 & 4{,}612 \\
C2 & fact & Recourse debt maturing in 2011         &   463 \\
C3 & fact & Recourse debt maturing in 2012         & 2{,}014 \\
C4 & fact & Recourse debt maturing in 2013         & 2{,}014 \\
C5 & fact & Recourse debt maturing in 2014         &   497 \\
C6 & fact & Recourse debt maturing in 2015         &   500 \\
C7 & fact & Recourse debt maturing after 2015      & 3{,}152 \\
C8 & formula & Percentage \(=\) (C7 \(/\) C1) \(\times 100\) & \ph \\
C9 & answer-interval & Percentage is in [68, 69]   & \ph \\
\bottomrule
\end{tabular}}
\caption{Claim catalog for the FinQA \emph{recourse debt} example.
Note that C3 and C4 share the suspicious value 2{,}014, exposing a
likely OCR / column-shift artefact in the original table.}
\label{tab:example_catalog}
\end{table}

%% file: tables/tab_example_market.tex
\begin{table}[!ht]
\centering
\setlength{\tabcolsep}{4pt}
\renewcommand{\arraystretch}{1.05}
\resizebox{0.30\textwidth}{!}{%
\begin{tabular}{lcrr}
\toprule
\textbf{ID} & \textbf{Status} & \textbf{Price} & \textbf{Confidence} \\
\midrule
C7  & accepted  & 1.00 & 1.00 \\
C8  & accepted  & 1.00 & 1.00 \\
C9  & accepted  & 1.00 & 0.99 \\
C1  & accepted  & 0.70 & 0.40 \\
C5  & uncertain & 0.50 & 0.00 \\
C2  & rejected  & 0.00 & 1.00 \\
C3  & rejected  & 0.00 & 1.00 \\
C4  & rejected  & 0.00 & 1.00 \\
C6  & rejected  & 0.00 & 1.00 \\
\bottomrule
\end{tabular}}
\caption{Cleared market for the FinQA \emph{recourse debt}
example. Eight of nine claims clear with confidence \(\ge 0.99\); only
C5 (the 2014 maturity) remains uncertain and is unused by the final
program.}
\label{tab:example_market}
\end{table}

%% file: custom.bib
@inproceedings{chen2021finqa,
  title={Finqa: A dataset of numerical reasoning over financial data},
  author={Chen, Zhiyu and Chen, Wenhu and Smiley, Charese and Shah, Sameena and Borova, Iana and Langdon, Dylan and Moussa, Reema and Beane, Matt and Huang, Ting-Hao and Routledge, Bryan R and others},
  booktitle={Proceedings of the 2021 Conference on Empirical Methods in Natural Language Processing},
  pages={3697--3711},
  year={2021}
}

@inproceedings{zhao2024docmath,
  title={DocMath-eval: Evaluating math reasoning capabilities of LLMs in understanding long and specialized documents},
  author={Zhao, Yilun and Long, Yitao and Liu, Hongjun and Kamoi, Ryo and Nan, Linyong and Chen, Lyuhao and Liu, Yixin and Tang, Xiangru and Zhang, Rui and Cohan, Arman},
  booktitle={Proceedings of the 62nd Annual Meeting of the Association for Computational Linguistics (Volume 1: Long Papers)},
  pages={16103--16120},
  year={2024}
}

@inproceedings{zhao2024financemath,
  title={Financemath: Knowledge-intensive math reasoning in finance domains},
  author={Zhao, Yilun and Liu, Hongjun and Long, Yitao and Zhang, Rui and Zhao, Chen and Cohan, Arman},
  booktitle={Proceedings of the 62nd Annual Meeting of the Association for Computational Linguistics (Volume 1: Long Papers)},
  pages={12841--12858},
  year={2024}
}

@inproceedings{cheng2022hitab,
  title={Hitab: A hierarchical table dataset for question answering and natural language generation},
  author={Cheng, Zhoujun and Dong, Haoyu and Wang, Zhiruo and Jia, Ran and Guo, Jiaqi and Gao, Yan and Han, Shi and Lou, Jian-Guang and Zhang, Dongmei},
  booktitle={Proceedings of the 60th Annual Meeting of the Association for Computational Linguistics (Volume 1: Long Papers)},
  pages={1094--1110},
  year={2022}
}

@inproceedings{zhao2022multihiertt,
  title={MultiHiertt: Numerical reasoning over multi hierarchical tabular and textual data},
  author={Zhao, Yilun and Li, Yunxiang and Li, Chenying and Zhang, Rui},
  booktitle={Proceedings of the 60th Annual Meeting of the Association for Computational Linguistics (Volume 1: Long Papers)},
  pages={6588--6600},
  year={2022}
}

@article{lu2023tabmwp,
  title={Dynamic prompt learning via policy gradient for semi-structured mathematical reasoning},
  author={Lu, Pan and Qiu, Liang and Chang, Kai-Wei and Wu, Ying Nian and Zhu, Song-Chun and Rajpurohit, Tanmay and Clark, Peter and Kalyan, Ashwin},
  journal={arXiv preprint arXiv:2209.14610},
  year={2022}
}

@inproceedings{pasupat2015wtq,
  title={Compositional semantic parsing on semi-structured tables},
  author={Pasupat, Panupong and Liang, Percy},
  booktitle={Proceedings of the 53rd Annual Meeting of the Association for Computational Linguistics and the 7th International Joint Conference on Natural Language Processing (Volume 1: Long Papers)},
  pages={1470--1480},
  year={2015}
}

@inproceedings{esgenius2025,
  title={Esgenius: Benchmarking llms on environmental, social, and governance (esg) and sustainability knowledge},
  author={He, Chaoyue and Zhou, Xin and Wu, Yi and Yu, Xinjia and Zhang, Yan and Zhang, Lei and Wang, Di and Lyu, Shengfei and Xu, Hong and Xiaoqiao, Wang and others},
  booktitle={Proceedings of the 2025 Conference on Empirical Methods in Natural Language Processing},
  pages={14623--14664},
  year={2025}
}

@article{hurst2024gpt,
  title={Gpt-4o system card},
  author={Hurst, Aaron and Lerer, Adam and Goucher, Adam P and Perelman, Adam and Ramesh, Aditya and Clark, Aidan and Ostrow, AJ and Welihinda, Akila and Hayes, Alan and Radford, Alec and others},
  journal={arXiv preprint arXiv:2410.21276},
  year={2024}
}

@article{liu2024deepseek,
  title={Deepseek-v3 technical report},
  author={Liu, Aixin and Feng, Bei and Xue, Bing and Wang, Bingxuan and Wu, Bochao and Lu, Chengda and Zhao, Chenggang and Deng, Chengqi and Zhang, Chenyu and Ruan, Chong and others},
  journal={arXiv preprint arXiv:2412.19437},
  year={2024}
}

@article{guo2025deepseek,
  title={Deepseek-r1: Incentivizing reasoning capability in llms via reinforcement learning},
  author={Guo, Daya and Yang, Dejian and Zhang, Haowei and Song, Junxiao and Wang, Peiyi and Zhu, Qihao and Xu, Runxin and Zhang, Ruoyu and Ma, Shirong and Bi, Xiao and others},
  journal={arXiv preprint arXiv:2501.12948},
  year={2025}
}

@article{liu2022tapex,
  title={TAPEX: Table pre-training via learning a neural SQL executor},
  author={Liu, Qian and Chen, Bei and Guo, Jiaqi and Ziyadi, Morteza and Lin, Zeqi and Chen, Weizhu and Lou, Jian-Guang},
  journal={arXiv preprint arXiv:2107.07653},
  year={2021}
}

@inproceedings{jiang2022omnitab,
  title={OmniTab: Pretraining with natural and synthetic data for few-shot table-based question answering},
  author={Jiang, Zhengbao and Mao, Yi and He, Pengcheng and Neubig, Graham and Chen, Weizhu},
  booktitle={Proceedings of the 2022 Conference of the North American Chapter of the Association for Computational Linguistics: Human Language Technologies},
  pages={932--942},
  year={2022}
}

@inproceedings{zhang2024tablellm,
  title={Tablellm: Enabling tabular data manipulation by llms in real office usage scenarios},
  author={Zhang, Xiaokang and Luo, Sijia and Zhang, Bohan and Ma, Zeyao and Zhang, Jing and Li, Yang and Li, Guanlin and Yao, Zijun and Xu, Kangli and Zhou, Jinchang and others},
  booktitle={Findings of the Association for Computational Linguistics: ACL 2025},
  pages={10315--10344},
  year={2025}
}

@article{li2024tablegpt2,
  title={Tablegpt2: A large multimodal model with tabular data integration},
  author={Su, Aofeng and Wang, Aowen and Ye, Chao and Zhou, Chen and Zhang, Ga and Chen, Gang and Zhu, Guangcheng and Wang, Haobo and Xu, Haokai and Chen, Hao and others},
  journal={arXiv preprint arXiv:2411.02059},
  year={2024}
}

@article{wang2025tabaf,
  title={General Table Question Answering via Answer-Formula Joint Generation},
  author={Wang, Zhongyuan and Zhang, Richong and Nie, Zhijie and Mao, Hangyu},
  journal={arXiv preprint arXiv:2503.12345},
  year={2025}
}

@inproceedings{wang2023napg,
  title={Napg: Non-autoregressive program generation for hybrid tabular-textual question answering},
  author={Zhang, Tengxun and Xu, Hongfei and van Genabith, Josef and Xiong, Deyi and Zan, Hongying},
  booktitle={CCF International Conference on Natural Language Processing and Chinese Computing},
  pages={591--603},
  year={2023},
  organization={Springer}
}

@article{lu2023chameleon,
  title={Chameleon: Plug-and-play compositional reasoning with large language models},
  author={Lu, Pan and Peng, Baolin and Cheng, Hao and Galley, Michel and Chang, Kai-Wei and Wu, Ying Nian and Zhu, Song-Chun and Gao, Jianfeng},
  journal={Advances in Neural Information Processing Systems},
  volume={36},
  pages={43447--43478},
  year={2023}
}

@inproceedings{wang2024chainoftable,
  title={Chain-of-table: Evolving tables in the reasoning chain for table understanding},
  author={Wang, Zilong Ryan and Zhang, Hao and Li, Chun-Liang and Eisenschlos, Julian M and Perot, Vincent and Wang, Zifeng and Miculicich, Lesly and Fujii, Yasuhisa and Shang, Jingbo and Lee, Chen-Yu and others},
  booktitle={International Conference on Learning Representations},
  volume={2024},
  pages={55587--55610},
  year={2024}
}

@inproceedings{sscot2024,
  title={Bridging neural and symbolic reasoning: A dual-system framework for interpretable question answering},
  author={Shi, Jihao and Ding, Xiao and Zhao, Hengwei and Liu, Ting and Qin, Bing},
  booktitle={ICASSP 2025-2025 IEEE International Conference on Acoustics, Speech and Signal Processing (ICASSP)},
  pages={1--5},
  year={2025},
  organization={IEEE}
}

@article{cao2025tablemaster,
  title={Tablemaster: A recipe to advance table understanding with language models},
  author={Cao, Lang and Liu, Hanbing},
  journal={arXiv preprint arXiv:2501.19378},
  year={2025}
}

@article{artemis2024,
  title={ARTEMIS-DA: an advanced reasoning and transformation engine for multi-step insight synthesis in data analytics},
  author={Hussain, Atin Sakkeer},
  journal={arXiv preprint arXiv:2412.14146},
  year={2024}
}

@inproceedings{gou2024critic,
  title={Critic: Large language models can self-correct with tool-interactive critiquing},
  author={Gou, Zhibin and Shao, Zhihong and Gong, Yeyun and Yang, Yujiu and Duan, Nan and Chen, Weizhu and others},
  booktitle={International Conference on Learning Representations},
  volume={2024},
  pages={57734--57811},
  year={2024}
}

@article{podoc2023,
  title={Tool documentation enables zero-shot tool-usage with large language models},
  author={Hsieh, Cheng-Yu and Chen, Si-An and Li, Chun-Liang and Fujii, Yasuhisa and Ratner, Alexander and Lee, Chen-Yu and Krishna, Ranjay and Pfister, Tomas},
  journal={arXiv preprint arXiv:2308.00675},
  year={2023}
}

@inproceedings{qian2023creator,
  title={Creator: Tool creation for disentangling abstract and concrete reasoning of large language models},
  author={Qian, Cheng and Han, Chi and Fung, Yi and Qin, Yujia and Liu, Zhiyuan and Ji, Heng},
  booktitle={Findings of the Association for Computational Linguistics: EMNLP 2023},
  pages={6922--6939},
  year={2023}
}

@article{cao2025fortune,
  title={Fortune: Formula-driven reinforcement learning for symbolic table reasoning in language models},
  author={Cao, Lang and Xu, Jingxian and Liu, Hanbing and Wang, Jinyu and Zhou, Mengyu and Dong, Haoyu and Han, Shi and Zhang, Dongmei},
  journal={arXiv preprint arXiv:2505.23667},
  year={2025}
}

@article{wei2022chain,
  title={Chain-of-thought prompting elicits reasoning in large language models},
  author={Wei, Jason and Wang, Xuezhi and Schuurmans, Dale and Bosma, Maarten and Xia, Fei and Chi, Ed and Le, Quoc V and Zhou, Denny and others},
  journal={Advances in neural information processing systems},
  volume={35},
  pages={24824--24837},
  year={2022}
}

@article{chen2022pot,
  title={Program of thoughts prompting: Disentangling computation from reasoning for numerical reasoning tasks},
  author={Chen, Wenhu and Ma, Xueguang and Wang, Xinyi and Cohen, William W},
  journal={arXiv preprint arXiv:2211.12588},
  year={2022}
}

@article{yao2023react,
  title={React: Synergizing reasoning and acting in language models},
  author={Yao, Shunyu and Zhao, Jeffrey and Yu, Dian and Du, Nan and Shafran, Izhak and Narasimhan, Karthik and Cao, Yuan},
  journal={arXiv preprint arXiv:2210.03629},
  year={2022}
}

@inproceedings{du2023improving,
  title={Improving factuality and reasoning in language models through multiagent debate},
  author={Du, Yilun and Li, Shuang and Torralba, Antonio and Tenenbaum, Joshua B and Mordatch, Igor},
  booktitle={Forty-first international conference on machine learning},
  year={2024}
}

@article{madaan2023selfrefine,
  title={Self-refine: Iterative refinement with self-feedback},
  author={Madaan, Aman and Tandon, Niket and Gupta, Prakhar and Hallinan, Skyler and Gao, Luyu and Wiegreffe, Sarah and Alon, Uri and Dziri, Nouha and Prabhumoye, Shrimai and Yang, Yiming and others},
  journal={Advances in neural information processing systems},
  volume={36},
  pages={46534--46594},
  year={2023}
}

@article{shinn2023reflexion,
  title={Reflexion: Language agents with verbal reinforcement learning},
  author={Shinn, Noah and Cassano, Federico and Gopinath, Ashwin and Narasimhan, Karthik and Yao, Shunyu},
  journal={Advances in neural information processing systems},
  volume={36},
  pages={8634--8652},
  year={2023}
}

@inproceedings{chen2022convfinqa,
  title={Convfinqa: Exploring the chain of numerical reasoning in conversational finance question answering},
  author={Chen, Zhiyu and Li, Shiyang and Smiley, Charese and Ma, Zhiqiang and Shah, Sameena and Wang, William Yang},
  booktitle={Proceedings of the 2022 conference on empirical methods in natural language processing},
  pages={6279--6292},
  year={2022}
}

@inproceedings{zhu2021tatqa,
  title={TAT-QA: A question answering benchmark on a hybrid of tabular and textual content in finance},
  author={Zhu, Fengbin and Lei, Wenqiang and Huang, Youcheng and Wang, Chao and Zhang, Shuo and Lv, Jiancheng and Feng, Fuli and Chua, Tat-Seng},
  booktitle={Proceedings of the 59th annual meeting of the Association for Computational Linguistics and the 11th international joint conference on natural language processing (volume 1: long papers)},
  pages={3277--3287},
  year={2021}
}

@inproceedings{reddy2024docfinqa,
  title={Docfinqa: A long-context financial reasoning dataset},
  author={Reddy, Varshini and Koncel-Kedziorski, Rik and Lai, Viet Dac and Krumdick, Michael and Lovering, Charles and Tanner, Chris},
  booktitle={Proceedings of the 62nd Annual Meeting of the Association for Computational Linguistics (Volume 2: Short Papers)},
  pages={445--458},
  year={2024}
}

@article{liu2025fino1,
  title={Fino1: On the transferability of reasoning-enhanced llms and reinforcement learning to finance},
  author={Qian, Lingfei and Zhou, Weipeng and Wang, Yan and Peng, Xueqing and Yi, Han and Zhao, Yilun and Huang, Jimin and Xie, Qianqian and Nie, Jian-yun},
  journal={arXiv preprint arXiv:2502.08127},
  year={2025}
}

@article{xie2024finben,
  title={Finben: A holistic financial benchmark for large language models},
  author={Xie, Qianqian and Han, Weiguang and Chen, Zhengyu and Xiang, Ruoyu and Zhang, Xiao and He, Yueru and Xiao, Mengxi and Li, Dong and Dai, Yongfu and Feng, Duanyu and others},
  journal={Advances in Neural Information Processing Systems},
  volume={37},
  pages={95716--95743},
  year={2024}
}

@misc{xie2023pixiu,
  title={Pixiu: A large language model, instruction data and evaluation benchmark for finance},
  author={Xie, Qianqian and Han, Weiguang and Zhang, Xiao and Lai, Yanzhao and Peng, Min and Lopez-Lira, Alejandro and Huang, Jimin},
  journal={arXiv preprint arXiv:2306.05443},
  year={2023}
}

@inproceedings{krumdick2024bizbench,
    title = "{B}iz{B}ench: A Quantitative Reasoning Benchmark for Business and Finance",
    author = "Krumdick, Michael  and
      Koncel-Kedziorski, Rik  and
      Lai, Viet Dac  and
      Reddy, Varshini  and
      Lovering, Charles  and
      Tanner, Chris",
    editor = "Ku, Lun-Wei  and
      Martins, Andre  and
      Srikumar, Vivek",
    booktitle = "Proceedings of the 62nd Annual Meeting of the Association for Computational Linguistics (Volume 1: Long Papers)",
    month = aug,
    year = "2024",
    address = "Bangkok, Thailand",
    publisher = "Association for Computational Linguistics",
    url = "https://aclanthology.org/2024.acl-long.452/",
    doi = "10.18653/v1/2024.acl-long.452",
    pages = "8309--8332"
}

@article{tan2025improved,
  title={Improved LLM agents for financial document question answering},
  author={Tan, Nelvin and Seng, Zian and Zhang, Liang and Shih, Yu-Ching and Yang, Dong and Salunkhe, Amol},
  journal={arXiv preprint arXiv:2506.08726},
  year={2025}
}

@inproceedings{wang2025tablemind,
author = {Jiang, Chuang and Cheng, Mingyue and Tao, Xiaoyu and Mao, Qingyang and Ouyang, Jie and Liu, Qi},
title = {TableMind: An Autonomous Programmatic Agent for Tool-Augmented Table Reasoning},
year = {2026},
isbn = {9798400722929},
publisher = {Association for Computing Machinery},
address = {New York, NY, USA},
url = {https://doi.org/10.1145/3773966.3777932},
doi = {10.1145/3773966.3777932},
booktitle = {Proceedings of the Nineteenth ACM International Conference on Web Search and Data Mining},
pages = {260–270},
numpages = {11},
location = {USA},
series = {WSDM '26}
}

@inproceedings{zhu2025financereasoning,
    title = "{F}inance{R}easoning: Benchmarking Financial Numerical Reasoning More Credible, Comprehensive and Challenging",
    author = "Tang, Zichen  and
      E, Haihong  and
      Ma, Ziyan  and
      He, Haoyang  and
      Liu, Jiacheng  and
      Yang, Zhongjun  and
      Rong, Zihua  and
      Li, Rongjin  and
      Ji, Kun  and
      Huang, Qing  and
      Hu, Xinyang  and
      Liu, Yang  and
      Zheng, Qianhe",
    editor = "Che, Wanxiang  and
      Nabende, Joyce  and
      Shutova, Ekaterina  and
      Pilehvar, Mohammad Taher",
    booktitle = "Proceedings of the 63rd Annual Meeting of the Association for Computational Linguistics (Volume 1: Long Papers)",
    month = jul,
    year = "2025",
    address = "Vienna, Austria",
    publisher = "Association for Computational Linguistics",
    url = "https://aclanthology.org/2025.acl-long.766/",
    doi = "10.18653/v1/2025.acl-long.766",
    pages = "15721--15749",
    ISBN = "979-8-89176-251-0"
}

@misc{yi2025reveal,
  title={ReVeal: Self-Evolving Code Agents via Reliable Self-Verification},
  author={Jin, Yiyang and Xu, Kunzhao and Li, Hang and Han, Xueting and Zhou, Yanmin and Li, Cheng and Bai, Jing},
  journal={arXiv preprint arXiv:2506.11442},
  year={2025}
}

@misc{arcs2025,
  title={ARCS: Agentic Retrieval-Augmented Code Synthesis with Iterative Refinement},
  author={Bhattarai, Manish and Cordova, Miguel and Vu, Minh and Santos, Javier and Boureima, Ismael and O'Malley, Dan},
  journal={arXiv preprint arXiv:2504.20434},
  year={2025}
}

@inproceedings{finagentcasc2025,
    title = "{D}avid vs. Goliath: Cost-Efficient Financial {QA} via Cascaded Multi-Agent Reasoning",
    author = "Liu, Chenghao  and
      Liu, Qian  and
      Zhu, Ziqin  and
      Fei, Hao  and
      Mahanti, Aniket",
    editor = "Christodoulopoulos, Christos  and
      Chakraborty, Tanmoy  and
      Rose, Carolyn  and
      Peng, Violet",
    booktitle = "Findings of the Association for Computational Linguistics: EMNLP 2025",
    month = nov,
    year = "2025",
    address = "Suzhou, China",
    publisher = "Association for Computational Linguistics",
    url = "https://aclanthology.org/2025.findings-emnlp.225/",
    doi = "10.18653/v1/2025.findings-emnlp.225",
    pages = "4212--4229",
    ISBN = "979-8-89176-335-7"
}

@misc{can-llms-debate2025,
  title={Can LLM Agents Really Debate? A Controlled Study of Multi-Agent Debate in Logical Reasoning},
  author={Wu, Haolun and Li, Zhenkun and Li, Lingyao},
  journal={arXiv preprint arXiv:2511.07784},
  year={2025}
}

@article{mad-fact2025,
  title={MAD-Fact: A Multi-Agent Debate Framework for Long-Form Factuality Evaluation in LLMs},
  author={Ning, Yucheng and Lin, Xixun and Fang, Fang and Cao, Yanan},
  journal={arXiv preprint arXiv:2510.22967},
  year={2025}
}

@article{mad-judge2025,
  title={Multi-agent debate for LLM judges with adaptive stability detection},
  author={Hu, Tianyu and Tan, Zhen and Wang, Song and Qu, Huaizhi and Chen, Tianlong},
  journal={Advances in Neural Information Processing Systems},
  volume={38},
  pages={46504--46540},
  year={2026}
}

@inproceedings{mad-math2025,
  title={Debate4MATH: Multi-agent debate for fine-grained reasoning in math},
  author={Zhang, Shaowei and Xiong, Deyi},
  booktitle={Findings of the Association for Computational Linguistics: ACL 2025},
  pages={16810--16824},
  year={2025}
}

@misc{deng2025tide,
  title={TIDE: Trajectory-based Diagnostic Evaluation of Test-Time Improvement in LLM Agents},
  author={Yan, Hang and Che, Xinyu and Xu, Fangzhi and Sun, Qiushi and Ding, Zichen and Cheng, Kanzhi and Zhang, Jian and Qin, Tao and Liu, Jun and Lin, Qika},
  journal={arXiv preprint arXiv:2602.02196},
  year={2026}
}

@article{zhang2024reactable,
author = {Zhang, Yunjia and Henkel, Jordan and Floratou, Avrilia and Cahoon, Joyce and Deep, Shaleen and Patel, Jignesh M.},
title = {ReAcTable: Enhancing ReAct for Table Question Answering},
year = {2024},
issue_date = {April 2024},
publisher = {VLDB Endowment},
volume = {17},
number = {8},
issn = {2150-8097},
url = {https://doi.org/10.14778/3659437.3659452},
doi = {10.14778/3659437.3659452},
journal = {Proc. VLDB Endow.},
month = apr,
pages = {1981–1994},
numpages = {14}
}

@article{xiao2024tradingagents,
  title={Tradingagents: Multi-agents llm financial trading framework},
  author={Xiao, Yijia and Sun, Edward and Luo, Di and Wang, Wei},
  journal={arXiv preprint arXiv:2412.20138},
  year={2024}
}

@article{shu2025finchart,
  title={Finchart-bench: Benchmarking financial chart comprehension in vision-language models},
  author={Shu, Dong and Yuan, Haoyang and Wang, Yuchen and Liu, Yanguang and Zhang, Huopu and Zhao, Haiyan and Du, Mengnan},
  journal={arXiv preprint arXiv:2507.14823},
  year={2025}
}

@article{yu2024fincon,
  title={Fincon: A synthesized llm multi-agent system with conceptual verbal reinforcement for enhanced financial decision making},
  author={Yu, Yangyang and Yao, Zhiyuan and Li, Haohang and Deng, Zhiyang and Jiang, Yuechen and Cao, Yupeng and Chen, Zhi and Suchow, Jordan W and Cui, Zhenyu and Liu, Rong and others},
  journal={Advances in Neural Information Processing Systems},
  volume={37},
  pages={137010--137045},
  year={2024}
}

@inproceedings{wang2026sheetbrain,
  title={SheetBrain: A neuro-symbolic agent for accurate reasoning over complex and large spreadsheets},
  author={Wang, Ziwei and Su, Jiayuan and Zhou, Mengyu and Zeng, Huaxing and Jia, Mengni and Lv, Xiao and Dong, Haoyu and Ma, Xiaojun and Han, Shi and Zhang, Dongmei},
  booktitle={Proceedings of the AAAI Conference on Artificial Intelligence},
  volume={40},
  number={40},
  pages={33800--33808},
  year={2026}
}
